\documentclass[sigconf,letterpaper,nonacm]{acmart}
\setcopyright{none}
\renewcommand\footnotetextcopyrightpermission[1]{}
\usepackage{siunitx}
\usepackage{bm}
\usepackage{rotating} 
\usepackage[acronym]{glossaries}
\makeglossaries
\usepackage{booktabs}
\usepackage{xspace}
\usepackage{subcaption}
\usepackage{comment}
\usepackage{todonotes}
\usepackage{graphicx}
\usepackage{tikz}
\usepackage{appendix}
\usepackage{ifthen}
\usepackage{csquotes}
\usepackage[inline]{enumitem}
\usepackage{mathtools}
\usepackage{cleveref}
\usepackage{url}
\usepackage{xcolor}    
\usepackage{soul}
\usepackage[nolist]{acronym}
\usepackage[most]{tcolorbox}
\usepackage{hyperref}

\usepackage{multirow}
\usepackage{pifont}

\AtBeginDocument{%
  }

\begin{document}
\newboolean{showcomments}
\setboolean{showcomments}{true} 
\newcommand\important[1]{\todo[inline]{\textbf{Important:} #1}}
\newcommand\gareth[1]{\textbf{\textcolor{red}{GT: #1}}}
\newcommand\Kiran[1]{\textbf{\textcolor{green}{KG: #1}}}
\newcommand\qm[1]{\todo[color=pink,inline]{\textbf{QM:} #1}}

\newcommand\summary[1]{
  \begin{tcolorbox}[%
    breakable,
    colback=white,
    colframe=black,
    sharp corners,
    boxrule=1pt,
    title=\textbf{Takehomes},
    titlerule=1pt,
    enhanced,
    skin first=enhanced,
    skin middle=enhanced,
    skin last=enhanced,
    left=1pt,    
    right=1pt,   
    boxsep=1pt
  ]
    #1
  \end{tcolorbox}
}

\newcommand\peixian[1]{\textbf{\textcolor{orange}{Peixian: #1}}}
\newcommand\qiming[1]{\textbf{\textcolor{pink}{Qiming: #1}}}
\newcommand\zf[1]{\textbf{\textcolor{green}{Zifan: #1}}}
\ifthenelse{\boolean{showcomments}} { }
{
	\renewcommand\important[1]{}
    \renewcommand\qiming[1]{}
    \renewcommand\zf[1]{}
    \renewcommand\peixian[1]{}
	\renewcommand\gareth[1]{}
    \renewcommand\Kiran[1]{}
}


\ifthenelse{\boolean{showcomments}}
{ \newcommand{\mynote}[3]{
		\protect\fbox{\bfseries\sffamily\scriptsize#1}
		{\small\textsf{\emph{\color{#3}{#2}}}}}}
{ \newcommand{\mynote}[3]{}}

\newcommand{\algvar}[1]{\ensuremath{\mathsf{#1}}}

\newcommand{\cf}{cf.\@\xspace}
\newcommand{\vs}{vs.\@\xspace}
\newcommand{\etc}{etc.\@\xspace}
\newcommand{\ala}{ala\@\xspace}
\newcommand{\wrt}{w.r.t.\@\xspace}
\newcommand{\etal}{\textit{et al.}\@\xspace}
\newcommand{\eg}{\textit{e.g.}\@\xspace}
\newcommand{\Eg}{\textit{E.g.}\@\xspace}
\newcommand{\ie}{\textit{i.e.}\@\xspace}
\newcommand{\Ie}{\textit{I.e.}\@\xspace}
\newcommand{\via}{\textit{via}\@\xspace}
\newcommand{\defacto}{\textit{de facto}\@\xspace}

\newcommand\para[1]{\vspace{0.05in} \noindent \textbf{#1}}
\newcommand{\pb}[1]{\vspace{0.75ex}\noindent{\bf \em #1}}
\newcommand{\mypara}[1]{\smallskip\noindent{\bf {#1}.}\xspace}

\newcommand\inlinesection[1]{{\bf #1.}}

\def\first{({\it i})\xspace}
\def\second{({\it ii})\xspace}
\def\third{({\it iii})\xspace}
\def\fourth{({\it iv})\xspace}
\def\fifth{({\it v})\xspace}
\def\sixth{({\it vi})\xspace}
\def\eg{\emph{e.g.,}\xspace}
\newcommand{\one}{({\em i})\xspace}
\newcommand{\two}{({\em ii})\xspace}
\newcommand{\three}{({\em iii})\xspace}
\newcommand{\four}{({\em iv})\xspace}
\newcommand{\five}{({\em v})\xspace}

\definecolor{verylightgray}{gray}{0.8}


\newcommand\vgap{\vskip 2ex}
\newcommand\marker{\vgap\ding{118}\xspace}

\def\na{--}
\def\unsure{?}
\def\missing{$!$}
\newcommand{\yes}{\ding{51}}
\newcommand{\no}{\ding{55}}
\DeclareRobustCommand\pie[1]{
\tikz[every node/.style={inner sep=0,outer sep=0, scale=1.5}]{
\node[minimum size=1.5ex] at (0,-1.5ex) {}; 
 \draw[fill=white] (0,-1.5ex) circle (0.75ex); \draw[fill=black] (0.75ex,-1.5ex) arc (0:#1:0.75ex); 
}
}
\def\L{\pie{0}} 
\def\M{\pie{-180}} 
\def\H{\pie{360}} 

\crefname{section}{Sec.}{Sec.}
\Crefname{section}{Section}{Sections}
\crefname{equation}{eq.}{eq.}
\crefname{figure}{Fig.}{Fig.s}
\Crefname{figure}{Figure}{Figures}

\floatstyle{plain}
\newfloat{lstfloat}{htbp}{lop}
\floatname{lstfloat}{Listing}
\def\lstfloatautorefname{Listing} 
\crefalias{lstfloat}{listing}

\DeclarePairedDelimiter\len{\lvert}{\rvert}
\DeclarePairedDelimiter\set{\{}{\}}

\begin{acronym}[Derp]
    \acro{aise}[LLM-SE]{LLM-based Search Engine}
    \acro{llm}[LLM]{Large Language Model}
    \acro{rag}[RAG]{Retrieval-Augmented Generation}
    \acro{seo}[SEO]{Search Engine Optimization}
    \acro{tse}[TSE]{Traditional Search Engine}
    \acro{cv}[CV]{Coefficient of Variation}
    \acro{mbfc}[MBFC]{Media Bias/Fact Check}
    \acro{lse}[LLM-SE]{LLM-based Search Engine}
    \acro{sd}[SD]{standard deviation}
    \acro{js}[JS]{Jaccard Similarity}
    \acro{wd}[WD]{Wasserstein Distance}
    \acro{rtd}[RTD]{Rank Turbulence Divergence}
    \acro{html}[HTML]{HyperText Markup Language}
    \acro{shap}[SHAP]{Shapley Additive exPlanation}
\end{acronym}

\title{Source Coverage and Citation Bias in LLM-based vs. Traditional Search Engines}

\author{Peixian Zhang}
\authornote{Both authors contributed equally to this research.} 
\affiliation{%
  \institution{The Hong Kong University of Science and Technology (Guangzhou)}
  \city{Guangzhou}
  \country{China}
}
\email{pzhang041@connect.hkust-gz.edu.cn}

\author{Qiming Ye}
\authornotemark[1] 
\affiliation{%
  \institution{The Hong Kong University of Science and Technology (Guangzhou)}
  \city{Guangzhou}
  \country{China}
}
\email{qiming@connect.hkust-gz.edu.cn}

\author{Zifan Peng}
\affiliation{%
  \institution{The Hong Kong University of Science and Technology (Guangzhou)}
  \city{Guangzhou}
  \country{China}
}
\email{zpengao@connect.hkust-gz.edu.cn}

\author{Kiran Garimella}
\affiliation{%
  \institution{Rutgers University}
  \city{New Brunswick}
  \country{United States}
}
\email{kiran.garimella@rutgers.edu}

\author{Gareth Tyson}
\affiliation{%
  \institution{The Hong Kong University of Science and Technology (Guangzhou)}
  \city{Guangzhou}
  \country{China}
}
\email{gtyson@ust.hk}

\begin{abstract}
\acp{aise} introduces a new paradigm for information seeking. Unlike \acp{tse} (\eg Google), these systems summarize results, often providing limited citation transparency. 
The implications of this shift remain largely unexplored, yet raises key questions regarding trust and transparency. 
In this paper, we present a large-scale empirical study of \acp{aise}, analyzing 55{,}936 queries and the corresponding search results across six \acp{aise} and two \acp{tse}.
We confirm that \acp{aise} cites domain resources with greater diversity than \acp{tse}.  
Indeed, \qty{37}{\percent} of domains are unique to \acp{aise}.
However, certain risks still persist: \acp{aise} do not outperform \acp{tse} in credibility, political neutrality and safety metrics.
Finally, to understand the selection criteria of \acp{aise}, we perform a feature-based analysis to identify key factors influencing source choice.  
Our findings provide actionable insights for end users, website owners, and developers.
\end{abstract}

\maketitle

\pagestyle{plain} 

\section{Introduction}
Search engines are one of the most frequently used services on the web.
They operate by crawling and indexing web content~\cite{google_search}, then ranking results based on things like relevance~\cite{tf_idf}, authority~\cite{hits}, reputation~\cite{pagerank} and experience~\cite{eeat}. 
For years, users have relied heavily on these rankings to access information~\cite{nielsen2016news,haim2018burst,trielli2019search}. 
However, the way users access and interact with information is changing with the rapid development of \acp{llm}~\cite{10.1145/3744746}.
Indeed, by 2027, it is projected that 90M American adults will use AI for search~\cite{genai_survey}.
Here, retrieval in \emph{LLM-based Search Engines} (LLM-SEs) involves an LLM agent conducting search activities on behalf of the user, then packaging up the results in a bespoke and easy-to-consume text summary.

However, to date, we know little about how this shift in information access will impact society at-large.
Technically, \acp{aise} diverge from \acp{tse} along two dimensions. 
First, \emph{Coverage} is supposedly enhanced through \ac{rag}~\cite{lewis2021retrievalaugmentedgenerationknowledgeintensivenlp}.  
\ac{rag} works by expanding queries into sub-queries~\cite{zhou2022least} and retrieving information from local knowledge database or online sources~\cite{yao2022react,trivedi2023interleaving,nakano2021webgpt,menick2022teaching}, reducing the chance of omissions. 
In contrast, users of \acp{tse} must typically perform those sub-queries themselves.
Second, \emph{Reliability} is expected to improve as \acp{llm} filter and re-rank citations with the reasoning capabilities of \acp{llm}~\cite{lewis2020retrieval}, allowing users to access key insights without inspecting multiple sources.

Although useful, this introduces multiple layers of indirection, potentially damaging transparency.
For example, prior studies indicate that \ac{rag} may produce unexpected bias~\cite{10.1145/3726302.3730230,dai2024bias}, and that \acp{llm} are prone to hallucinations~\cite{huang2025survey} or even harmful content generation~\cite{giorgi2025human}.
Researchers have found that participants who use \acp{aise} are unaware of errors in the responses~\cite{spatharioti2025effects}, and that more malicious sources appear in \acp{aise} than \acp{tse}~\cite{luo2025unsafe}.
The existing solution to these challenges is for \acp{aise} to include source citations~\cite{gao-etal-2023-enabling}, offering external URLs that evidence the assertions made in the response. 
This allows users to follow such URLs, and check the correctness of the LLM summaries. 
However, to date, we lack a systematic evaluation of how \acp{aise} use these citations, nor how they differ to the results offered by \acp{tse}.

Therefore, this paper explores the source citation of \acp{aise}, as well as their ability to deliver information transparency and reliability compared to \acp{tse}.
We evaluate six different \acp{aise}, benchmarking the results they return against to two major \acp{tse} (Google and Bing). To the best of our knowledge, this is the first study to perform a comprehensive evaluation across a wide range of \acp{aise}, offering insights into their sourcing behavior, reliability, and biases.
We explore the following research questions:

\begin{itemize}
    \item \textbf{RQ1:} What are the sources cited by \acp{aise}, and how do they differ from those returned by \acp{tse}?
    
    \item \textbf{RQ2:} How is the quality of sources cited by \acp{aise}, and how does it compare to \acp{tse}?
    
    \item \textbf{RQ3:} Which features influence the likelihood of a website being cited by \acp{aise} compared to \acp{tse}?
\end{itemize}

Our contributions are as follows:
\begin{enumerate}
    \item We present the first large-scale measurement study comparing Google and Bing with six popular \acp{aise}. We gather a dataset of 55,936 queries, covering 124{,}287 unique domains and 1{,}418{,}733 unique citation hyperlinks. We will make this available for research use.
   
    \item Beyond returning fewer sources, \acp{aise} diverge from \acp{tse} in content selection: \qty{37}{\percent} of domains in \ac{aise} results are absent from \ac{tse} outputs. \acp{aise} exhibit a less concentrated distribution and favor domains with lower user popularity.

    \item Gemini returns frequent far-left citations, while Grok has the lowest partisan neutrality; both rely on limited sources that further limit overall reliability. Besides, \acp{aise} do not outperform \acp{tse} in cyber safety despite lower source volumes.

    \item Feature-based analysis shows that domains favored by \acp{aise} generally exhibit more structured, hierarchical HTML, easier-to-read text, lower domain popularity, and more outlinks to reputable sources compared to \acp{tse}.
\end{enumerate}

\section{Background \& Related Work}

\pb{\acp{tse}}
\acp{tse} (\eg Google~\cite{google_search}, Bing~\cite{bing_search}) use crawlers to discover pages via links and sitemaps~\cite{sitemap}. Indexers then build inverted indexes (\eg TF‑IDF~\cite{tf_idf}) with term frequencies and positions. Finally, ranking combines link analysis (\eg PageRank~\cite{pagerank}) and quality criteria (\eg E‑E‑A‑T~\cite{eeat}).
The final output for the user is presented as a ranked list of links.
These ranking positions shape user impressions and direct traffic to the linked domains~\cite{guan2007eye,pan2007google}.
Though accounting for over half of global web traffic~\cite{seo_traffic}, \acp{tse} also raise concerns.
For example, users may blindly trust top-ranked results or well-known domains, reinforcing rank bias~\cite{wang2016learning} and authority-domain bias~\cite{ieong2012domain}.
The prevalence of highly similar information can further reinforce echo chambers~\cite{diaz2022echo}, which accelerates political polarization and introduces additional bias~\cite{poudel2025social}.

\pb{\acp{aise}}
In contrast, \acp{aise} adopt a fundamentally different approach by integrating retrieval with language generation through the \ac{rag} paradigm~\cite{lewis2021retrievalaugmentedgenerationknowledgeintensivenlp}. 
Following the Least-to-Most Prompting framework~\cite{zhou2022least}, they first interpret user intent and decompose complex queries into sub-queries. 
These sub-queries are then used to retrieve relevant content from either local knowledge bases or online sources~\cite{yao2022react,trivedi2023interleaving}.  
For queries requiring recent or specialized information (\eg news), \acp{aise} additionally perform real-time web retrieval~\cite{nakano2021webgpt,menick2022teaching}. 
The retriever then ranks and selects the most relevant texts for downstream processing.
The \ac{llm} generator then interprets these documents, synthesizes the retrieved content~\cite{lewis2020retrieval}, and produces a concise natural-language summary with inline citations~\cite{gao-etal-2023-enabling}.  
However, LLMs themselves may exhibit biases. 
For example, misinformation in retrieved documents can propagate into generated answers~\cite{deng2025cram}.
Prior work also demonstrates that \acp{llm} exhibit political bias when rating news sources~\cite{yang2025accuracy} and may generate malicious links~\cite{luo2025unsafe}.
Additionally, results from Search Arena~\cite{miroyan2025searcharenaanalyzingsearchaugmented} show that users are strongly influenced by the number and type of citations, even when these attributions are incorrect.

\pb{Comparison of \acp{tse} vs. \acp{aise}} 
By shifting from link ranking to answer generation, \acp{aise} mark a fundamental departure from the click-driven exploration of traditional search. 
Prior work shows that while \acp{aise} can outperform \acp{tse} on complex, decision-oriented tasks~\cite{spatharioti2025effects}, users also exhibit over-reliance on incorrect outputs~\cite{kabir2023answers}.
In contrast, \cite{wazzan2024comparing} demonstrate that \acp{tse} users achieve higher accuracy in an image localization task, highlighting task-dependence in performance. Citations have been shown to increase user trust in \acp{aise} responses~\cite{ding2025citations}, and topic analyses indicate that users prefer \acp{aise} for semantically rich, language-intensive tasks~\cite{caramancion2024large}. 
However, \acp{aise} also exhibit higher rates of unsafe or malicious link generation compared to \acp{tse}~\cite{luo2025unsafe}. 
Together, these findings illuminate how users interact with \acp{aise}. Yet, they do not study the sources that \acp{aise} cite, nor how they differ from \acp{tse}.
Thus, in this work, we address this gap by conducting the first large-scale analysis of \acp{aise} citation patterns.
We simulate the search activities on both \acp{tse} and \acp{aise}, focusing on their sourcing patterns and the implications for information exposure.
\section{Data Collection and Annotation}
\label{sec:data_collection_and_annotation}

\begin{figure}[t!]
    \centering
    \includegraphics[width=\linewidth]{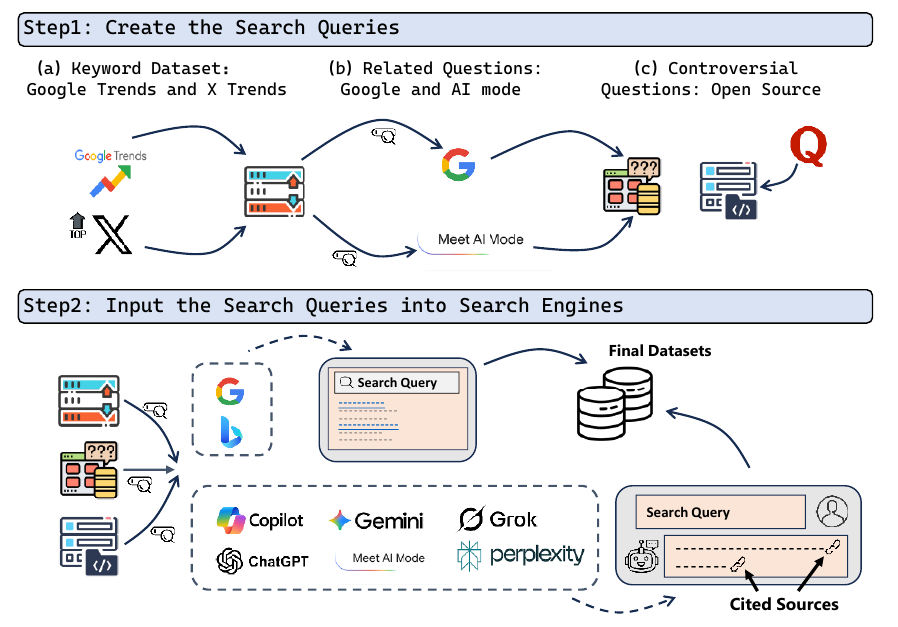}
    \caption{
    The Pipeline of Data Collection} 
    \label{fig:data_pipeline}
\end{figure}

We begin by introducing our data collection and annotation methodologies. 
Figure~\ref{fig:data_pipeline} presents the overall pipeline of our data collection work.
Note that \acp{aise} may process multiple pages before presenting the selected sources to the user. 
However, we focus exclusively on user-facing citations to capture the end-user experience.
We capture a snapshot of popular queries to compare the sources returned by search engines.

\subsection{Search Engines}
We define an \ac{aise} as an LLM system that can retrieve web content and provide cited answers paired with summarized text.
We therefore first identify publicly available \acp{llm} that provide web search functionality. 
We include 6 widely used platforms: \textit{ChatGPT}, \textit{Gemini}, \textit{Perplexity}, \textit{Grok}, \textit{Google AI Mode} (short as AI Mode), and \textit{Copilot Search in Bing} (short as Copilot).
For comparison, we also include \textit{Google} and \textit{Bing} as representatives for \acp{tse}.

\subsection{Search Query Corpus}
Prior work~\cite{miroyan2025searcharenaanalyzingsearchaugmented,yang-etal-2018-hotpotqa} shows that \acp{tse} favor keyword-based queries, whereas users express real-world factual intents in natural language on \acp{aise}. 
Therefore, we augment the query corpus with keywords, related questions, and controversial queries.

\pb{Keyword Queries.}
We start by collecting a seed keyword dataset to represent search activities.
Here, we use \textit{Google Trends} and \textit{X (formerly Twitter)} (Step 1(a) in \Cref{fig:data_pipeline}). 
These trending keywords serve as representative inputs, reflecting what users actively search for across social media and search engines.
Both of them are representative of real-time user interests and are widely used in the prior research~\cite{annamoradnejad2019comprehensive,jun2018ten}.
We download Google Trends\footnote{\url{https://trends.google.com/trends/}} and X Trends by trends24\footnote{\url{https://trends24.in/united-states/}} everyday between Jul. 16 to Aug. 10 2025 across the USA.
In total, we collect 7{,}519 unique keywords from Google Trends and 5{,}593 from X, and we include the search results corresponding to these keywords in our dataset.  

\pb{Related Question Query.}
Relying solely on single keywords is insufficient, as they may be ambiguous. 
Prior studies have shown that users interacting with \acp{aise} tend to prefer sentence-style queries rather than isolated keywords~\cite{caramancion2024large,luo2025unsafe,wazzan2024comparing}.
To align with this behavior, we generate natural sentence queries from the above trending keyword lists. As illustrated in \Cref{fig:data_pipeline} Step 1(b), we take the keywords generated from Step 1(a), and search for them using both Google Search and Google AI mode.
As well as returning the search results, these two services also return a set of recommended related search queries. These are generated from prior real searches, and give insight into typical queries launched by users.
This approach ensures that our query set reflects realistic user input patterns.
In total, this process yields 37{,}931 unique questions from Google search result and 2{,}612 unique questions from AI mode.

\pb{Controversial Questions.}
To complement the related questions, we include an existing controversial query dataset to examine how search engines handle sensitive or polarizing content (Step 1(c) in \Cref{fig:data_pipeline}).  
We utilize an open-source dataset containing sentence-level queries sourced from popular Quora questions that have been identified as controversial by large language models (LLMs)~\cite{sun2023Delphi}.
This dataset covers 2{,}281 queries. These queries are curated to reflect socially and politically sensitive topics, enabling us to assess how different \acp{aise} handle nuanced or polarizing content.

\subsection{Search Query and Response Processing} 
Upon completing the construction of the search query dataset, we initiate the search process as Step 2 by inputting all keywords and sentence queries to both the \acp{tse} and \acp{aise}. 
We then employ DrissionPage\footnote{a web automation tool based on Python~\cite{drissionPage}} to simulate user interactions with browser.
Note, URLs may appear multiple times within a single response; however, we count each only once.

\pb{\acp{tse}.}
To minimize personalization effects from \acp{tse}~\cite{hannak2013measuring}, we configure the browser to operate in incognito mode. 
Previous work shows that the click-through rate for the second page is around \qty{0.63}\% on Google~\cite{dean2025ctr}, thus we focus on first-page results to reflect the content most end users encounter~\cite{gezici2021evaluation}.
We explicitly flag URLs marked with ``Ad'' or ``Sponsored'' tags as advertisements, while preserving their original positions in the ranked list for completeness.  
In total, our dataset contains 481{,}565 unique URLs from Google and 218{,}122 unique URLs from Bing.

\pb{\acp{aise}.}
For \acp{aise}, we collect data based on browser-based search mode.\footnote{Note, some \acp{aise} (\eg ChatGPT and Gemini) provide the desktop version.}
We operate the browser in incognito mode to minimize potential personalization effects. 
\Cref{tab:summay_aise} details the additional actions we performed for each \acp{aise}.
Notably, ChatGPT is the only \ac{aise} that mandates user authentication; thus, we configure it to avoid referencing prior interactions and disable memory features to ensure consistency across sessions.
We then parse the full response payloads and extract the returned items and citations as the cited sources, as illustrated in Step~2 of \Cref{fig:data_pipeline}.

\begin{table}[t]
\caption{The Summary of AI Search Engines.}
\label{tab:summay_aise}
\resizebox{\columnwidth}{!}{
\begin{tabular}{@{}llllll@{}}
\toprule
\textbf{Name}  & \textbf{Version}     & \textbf{Login} & \textbf{Search Button} & \textbf{\# of Cited Sources} &  \\ \midrule
AI Mode    & Gemini-based*    & \ding{55} & \ding{55}        & 414,524             &  \\
ChatGPT    & gpt-5.0-instant     & \ding{51} & \ding{51}        & 206,590               &  \\
Copilot    & GPT-based*    & \ding{55} & \ding{55}        & 1,029,015              &  \\
Gemini     & Gemini-2.5-flash  & \ding{55} & \ding{55}        & 182,541          &  \\
Grok       & Grok 3.0 & \ding{55} & \ding{55} & 
62,420       &  \\
Perplexity & Sonar   & \ding{55} & \ding{51}        & 280,699           &  \\ \bottomrule
\multicolumn{5}{@{}l}{\footnotesize AI Mode and Copilot do not publish exact version numbers.}\\
\multicolumn{5}{@{}l}{\footnotesize For clarity, our analysis exclusively uses Copilot in Bing.}
\end{tabular}}
\end{table}

\subsection{Data Annotations}
\label{sec:data_annotation}
\pb{Search Query Annotation.}
We annotate each query according to Google Trends' categorization scheme (see \Cref{app:query_categorization}).
To this end, we employ a large language model (LLM) as a classifier.
The detailed methodology and validation used for category prediction are provided in \Cref{app:query_categorization}.
Overall, the query topics are broadly distributed across 26 categories.
The most represented categories are \textit{Arts \& Entertainment} (\qty{27}{\percent}), \textit{People \& Society} (\qty{19}{\percent}), and \textit{Sports} (\qty{17}{\percent}), confirming strong topical diversity.

\pb{Cited Domains.}
To categorize the cited websites, we employ the Advanced Classification Engine (ACE) from Forcepoint~\cite{forcepoint}.
ACE performs URL categorization and is used in prior studies~\cite{vallina2020mis}.
To evaluate source popularity, we employ three widely used third-party datasets.
To quantify the number of visits a website receives, we use the Tranco Ranking~\cite{tranco_reputation}.
To assess source quality, we incorporate two widely used third-party datasets: Media Bias/Fact Check (MBFC)~\cite{mdfc} for credibility and political bias, and VirusTotal~\cite{virustotal} for security and malicious-domain detection.
Together, these tools provide domain-level annotations on categorization~\cite{bouwman2022helping}, popularity~\cite{10.5555/3359012.3359022}, political bias~\cite{10.1145/3711542.3711601}, and cyber security risks~\cite{imc_virus_total_threshold}.
Note, not all domains cited by the search engines can be matched to the above datasets; for example, MBFC only covers news websites. 
The detailed coverage rate and validation results are provided in \Cref{app:annotation_results}.

\section{Source Citation Behaviors (RQ1)}
\label{sec:rq1}
Unlike \acp{tse}, which typically return a ranked list of hyperlinks, \acp{aise} provide a summarized response supported by fewer selected URL sources.
However, it is unclear whether \acp{aise} differ the information landscape with well-selected sources. 
To address this gap, we measure the sourcing behaviors beginning.

\subsection{Comparing Citation Patterns}  
\label{sec:number_of_sources}
We begin by analyzing the distribution of sources cited in the search results. 
This allows us to assess how different engines prioritize, select, and present reference sources.

\begin{figure}[t]
    \centering
    \begin{subfigure}[b]{0.48\columnwidth}
        \centering        \includegraphics[width=\textwidth]{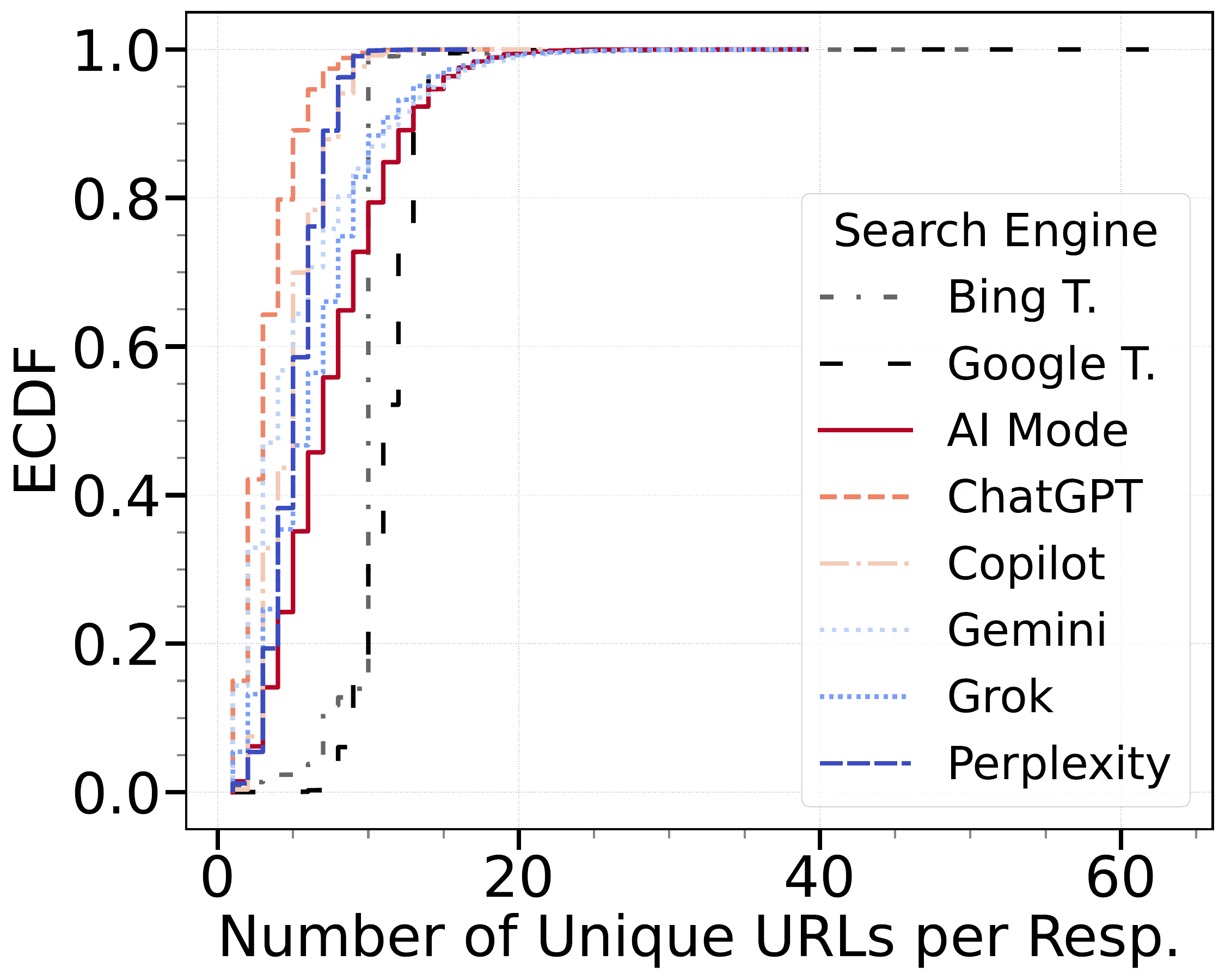}
        \caption{}
        \label{fig:url_ecdf}
    \end{subfigure}
    \hfill
    \begin{subfigure}[b]{0.48\columnwidth}
        \centering
    \includegraphics[width=\textwidth]{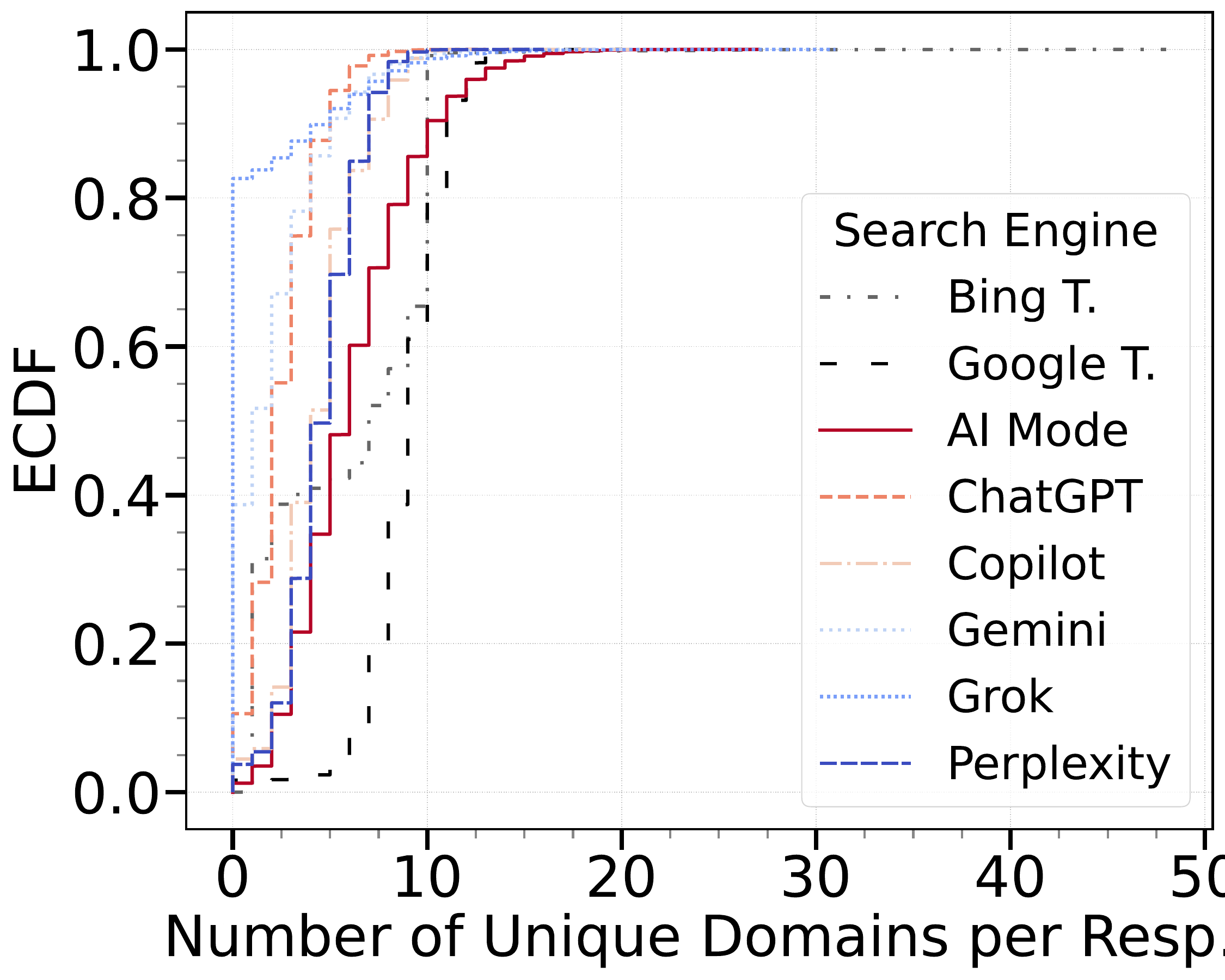}
        \caption{}
        \label{fig:domain_ecdf}
    \end{subfigure}
    \caption{ECDFs: number of unique (a) websites and (b) domains per response across search engines.}
    \label{fig:url_domain_ecdfs}
\end{figure}

\pb{Number of Sources.} 
We first assess the number of distinct sources embedded per response.  
\Cref{fig:url_ecdf} and \Cref{fig:domain_ecdf} illustrate the distributions of unique URLs and domains.\footnote{URLs or domains may appear multiple times within different parts of the response text due to repeated embedding by the search engines.}
As expected, all \acp{aise} source fewer URLs (mean: 4.3) and domains (mean: 3.4) compared to \acp{tse} (mean: 10.3 URLs and 7.3 domains).
Across all \acp{aise}, fewer than ten distinct URLs appear in 80\% of responses.
Notably, Grok and Gemini are the least likely to cite external sources, with 82\% and 38\% of their responses containing no cited websites, respectively. 
This behavior may stem from a stronger reliance on internal knowledge bases or stricter display thresholds, causing these engines to prioritize model-generated content over external citations. 
However, prior work shows that limited sourcing can undermine user trust in search systems~\cite{miroyan2025searcharenaanalyzingsearchaugmented}.

\begin{figure}[t]
    \centering
    \centering
    \includegraphics[width=0.5\linewidth]{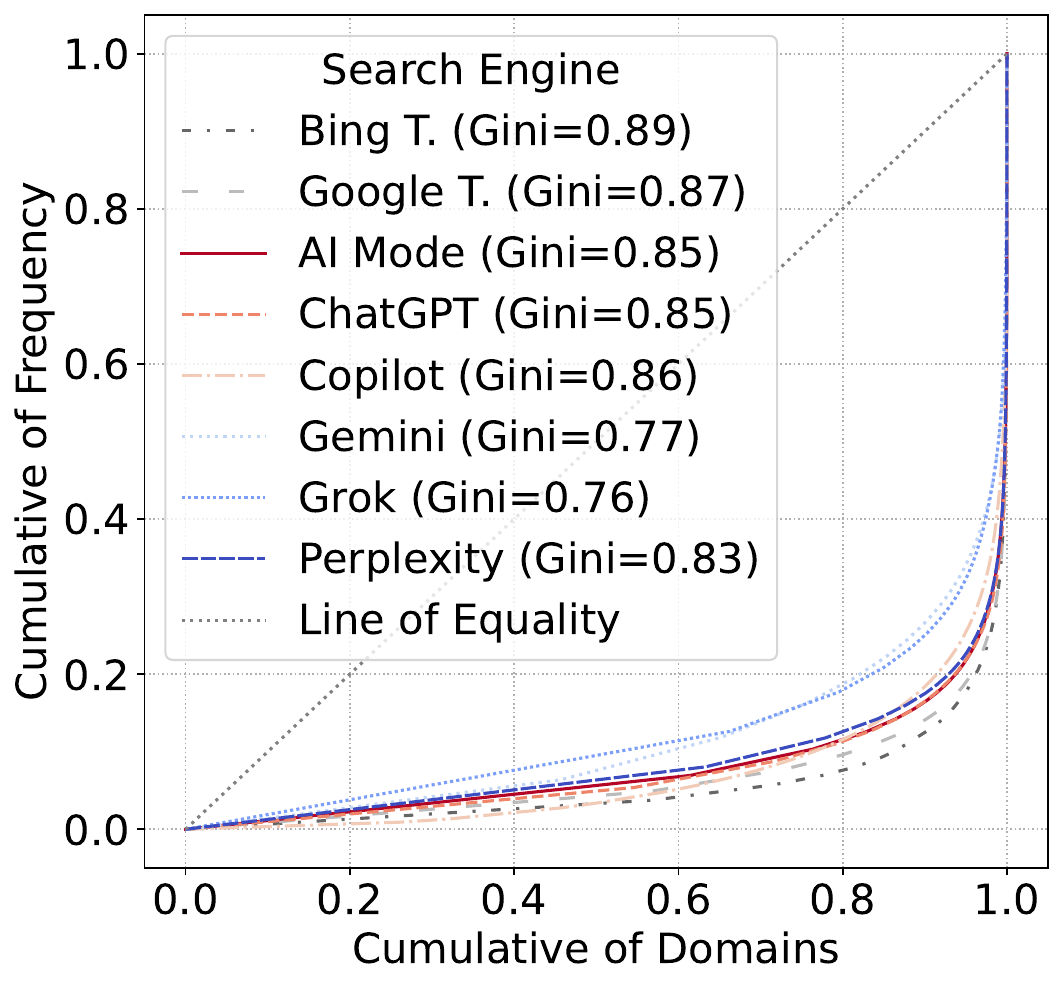}
    \caption{Lorenz curve illustrating domain frequency distribution and corresponding Gini index.}
    \label{fig:gini_index}
\end{figure} 

\pb{Domain Concentration.}
Source concentration is a central concern in prior studies of \acp{tse}~\cite{diaz2022echo}, and similar risks have been identified in LLMs, where authority bias shapes how models evaluate content~\cite{chen2024humans}. 
We therefore use the Gini index to quantify domain concentration for each search engine, capturing the extent to which they disproportionately return certain domains. 
A higher Gini index ($\to 1$) indicates that a few domains dominate the citations, whereas a lower value ($\to 0$) suggests an even distribution.

\Cref{fig:gini_index} presents the Lorenz Curve~\cite{gastwirth1971general} for each search engine.
The results show that all \acp{aise} exhibit lower Gini indices than \acp{tse}, indicating that \acp{aise} return a slightly more uniform spread of domains.
To validate this, we perform a statistical test for each system pair (\ie \acp{aise} vs.\ \acp{tse}) following the methodology of \cite{xu2000inference} (details in \Cref{app:gini}).
Our results show that Gemini, Grok and ChatGPT differ significantly from more than one \ac{tse}. 
ChatGPT diverges only from Google ($p=0.026$), likely reflecting its reliance on Bing, whereas Gemini and Grok differ significantly from both Bing and Google ($p<0.001$). 
Grok reports using no third-party provider, while Gemini depends on Google, and we conjecture that their divergence may also stem from their lower citation counts (see \S\ref{fig:url_domain_ecdfs}). 
Although Gemini and Grok exhibit lower Gini indices than the \acp{tse}, they nonetheless return fewer sources overall. 
However, the extent to which such source diversity impacts the actual user retrieval experience remains an open question.
This motivate us to measure the source quality in \S\ref{sec:rq2}.

\begin{figure}
    \centering
    \includegraphics[width=0.85\linewidth]{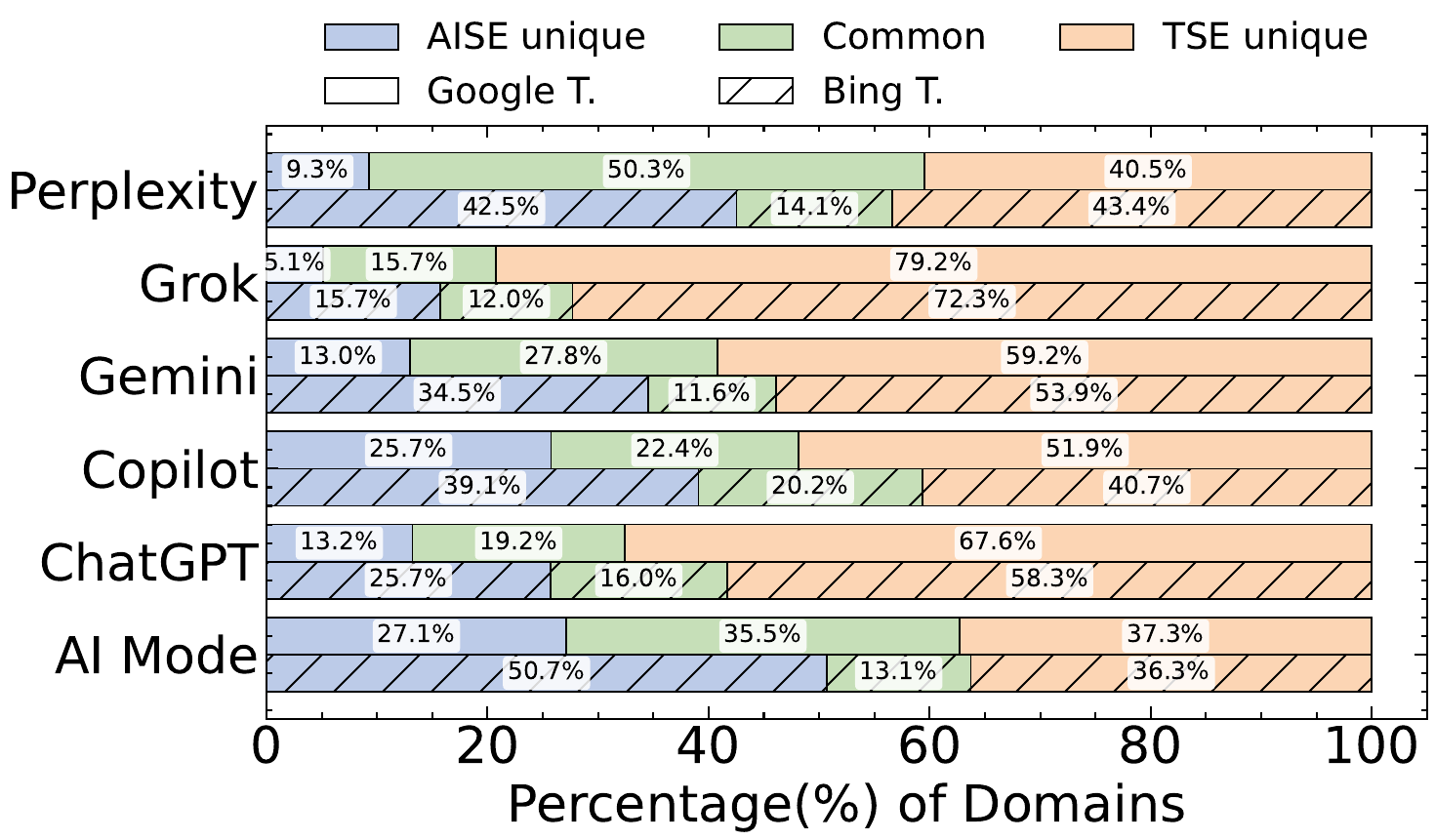}
    \caption{
    Percentage of domain overlap:
    \emph{Unique} indicates domains appearing in only one search engine type, while \emph{Common} indicates sharing between \acp{aise} and \acp{tse}.
    }
    \label{fig:unique_domains}
\end{figure}

\subsection{Disparity in Domain Selection}
\label{sec:disparity}

We next assess the extent to which each search engine returns unique domains. 
High distinctiveness indicates that users may retrieve varied information depending on the search engine.

\pb{Overall Disparity.}
\Cref{fig:unique_domains} presents the fraction of domains that are unique to \acp{aise}, unique to \acp{tse}, or overlapping between the two. 
Here, \emph{Unique} refers to domains appearing exclusively in the corresponding \acp{aise} or \acp{tse}, whereas \emph{Common} denotes domains shared between both.
Surprisingly, large differences are observed between \acp{aise} and \acp{tse}: only 38\% of domains appear in both (\ie common), whereas 37\% are unique to \acp{aise} results.
Similar patterns also persist at the response level, with an average overlap of less than 40\% between any \acp{aise} and \acp{tse} pair (see \Cref{app:coverage_ratio}).
Even for services provided by the same company (\eg Google, AI Mode, and Gemini), the sourced domains differ substantially ($\geq30\%$). 
This divergence likely arises because the \acp{aise} decomposes the original queries rather than using the full user-provided context when retrieving content from external sources~\cite{press2023measuring}.
This confirms that users relying solely on \acp{aise} are exposed to a distinct set of information sources compared to those using \acp{tse} alone.

\begin{figure}[t] 
  \centering
  \hfill
  \begin{subfigure}[t]{\linewidth}
    \centering
    \includegraphics[width=0.75\linewidth]{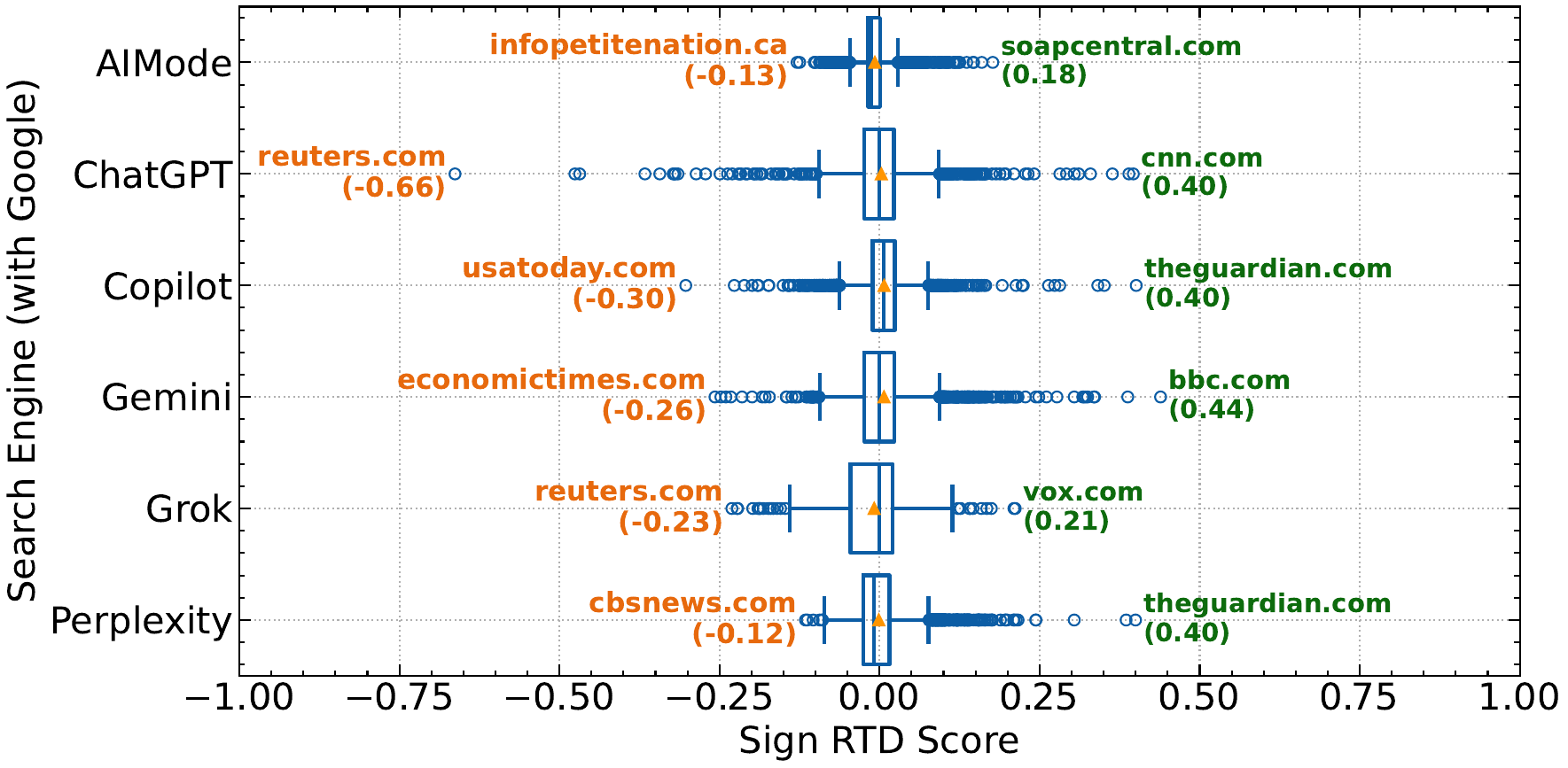}
    \caption{With Google}
    \label{fig:rtd_withgoogle}
  \end{subfigure}
  \hfill
  \begin{subfigure}[t]{\linewidth}
    \centering
    \includegraphics[width=0.75\linewidth]{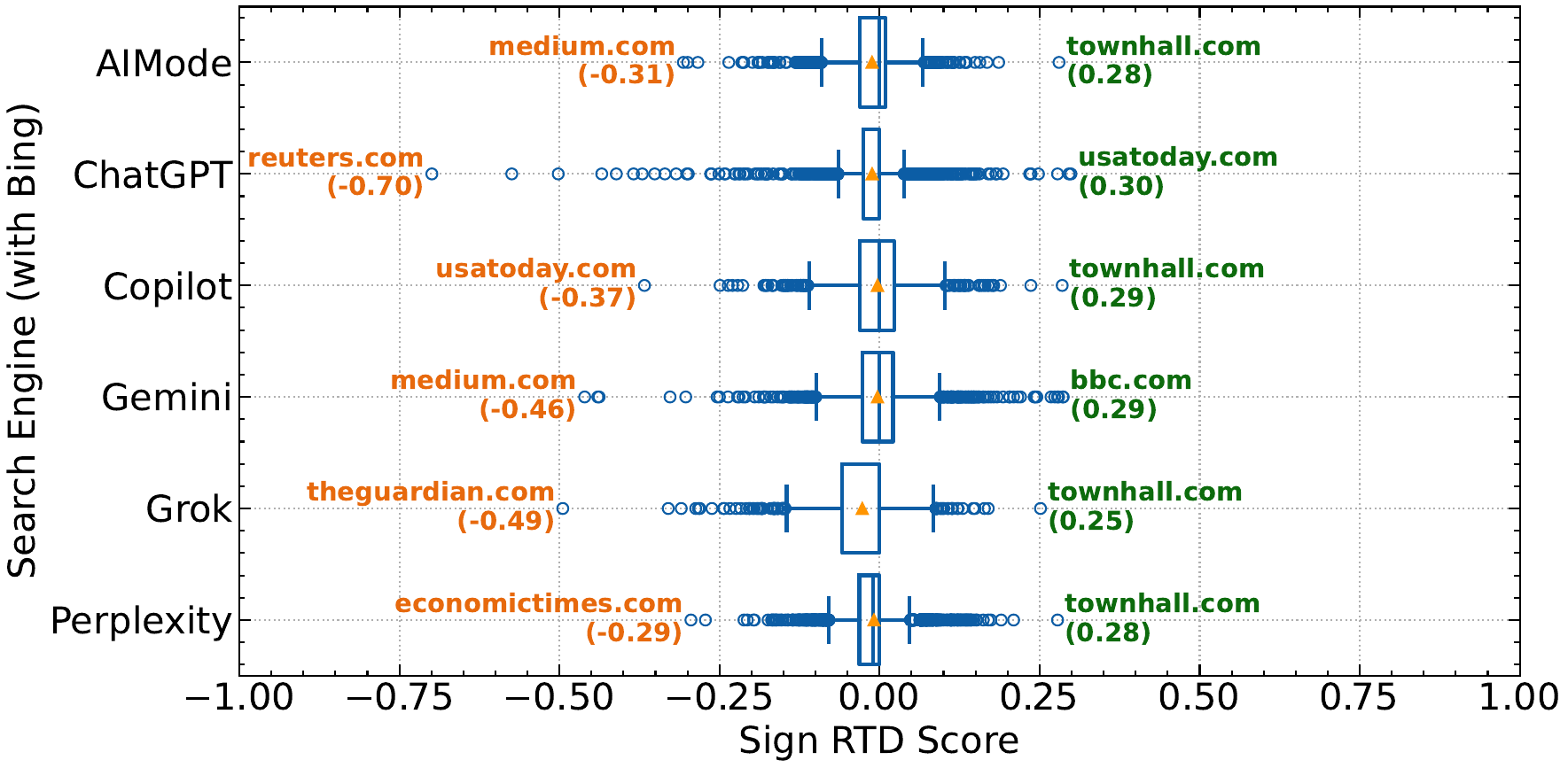}
    \caption{With Bing}
    \label{fig:rtd_withbing}
  \end{subfigure}
  \caption{RTD of news relevant domains distribution between \acp{aise} and (a) Google (b) Bing. The \textcolor{orange}{domains} on the left are more likely to appear in \acp{aise} response and \textcolor{green}{domains} on the right are more likely to appear in \acp{tse} response.}
  \label{fig:rtd}
\end{figure} 

\pb{Disparity in News Sources.} 
Although \acp{aise} diverge from \acp{tse} in their overall domain distribution, it remains unclear whether they favor specific sources. 
Such preferences are important, as over-reliance on a single source can amplify biases and reduce diversity, a concern particularly relevant for news content.
To investigate this, we extract all news domains (as labeled by VirusTotal, see \S\ref{sec:data_annotation}) from the search results. 
This results in 2{,}675 unique domains returned by both \acp{aise} and \acp{tse}. 
We then employ Rank Turbulence Divergence (RTD)~\cite{dodds2023allotaxonometry, poudel2024navigating} to quantify differences between \acp{aise} and \acp{tse} results (see \Cref{app:rtd} for details). 
RTD measures domain-specific preference: a score close to $1$ indicates strong presence in \ac{tse} results, whereas a score near $-1$ signals a higher likelihood of appearing in \acp{aise}.
\Cref{fig:rtd} presents the \ac{rtd} distributions across domains for \acp{aise} compared with (a) Google and (b) Bing.  
Each subplot highlights the top-ranked domains with the highest \ac{rtd} values cited by the respective engines.
Although the overall difference centers near zero, certain domains receive disproportionate emphasis.
As an illustrative case, ChatGPT more frequently surfaces the original news agencies Reuters and AP News (Google RTD = –0.48, Bing RTD = –0.41). In contrast, \acp{tse} favor established broadcasters (CNN, USA TODAY) and opinion-based sources (Townhall). 
Since news sources often have established political positions, such divergence could introduce bias, which we further investigate in \S\ref{sec:source_reliability}.

\subsection{Measuring Source Popularity}
\label{sec:source_popularity}

Finally, the \emph{popularity} of domains is measured through two metrics, as it can directly influence the perceived authority of a search engine:
\begin{enumerate*}
    \item ranking positions on Google and Bing, where higher placements indicate greater search-driven exposure~\cite{seo_traffic};\footnote{Note that organic search via \acp{tse} still accounts for over half of global web traffic as of 2025~\cite{seo_traffic}.}
    \item Tranco rankings~\cite{tranco}, providing a platform-agnostic estimate of web traffic for each domain. 
\end{enumerate*}
Together, these metrics capture both the likelihood of domains being discovered and visited, serving as a proxy for domain popularity and its impact on perceived search authority.

\begin{figure}[t]
    \centering
    \includegraphics[width=0.9\columnwidth]{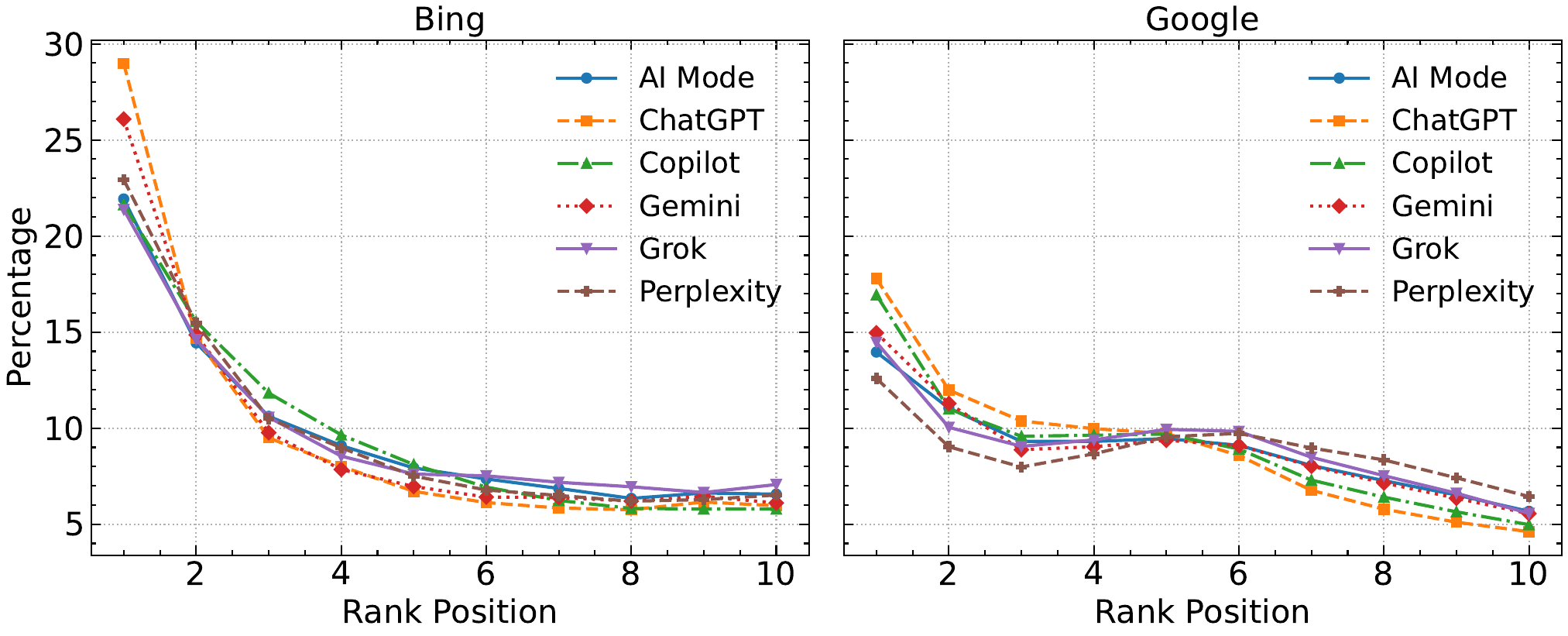}  
    \caption{The figure shows where overlapping domains between \acp{aise} and \acp{tse} appear in each search. It aggregates across all searches to report the frequency of these domains' ranks in \acp{tse}.}
    \label{fig:domain_rank}
\end{figure}

\pb{Search-driven Exposure.}
We begin by assessing the rank positions of the domains that appear in both the \ac{aise} responses and the top \ac{tse} results. 
For each search query, we identify all domains that overlap between \acp{aise} and \acp{tse} in a pairwise comparison. 
This accounts for only 37\% of domains in our dataset (see \S\ref{sec:number_of_sources}).
If a domain appears at multiple ranks within \acp{tse} results for a single search, all instances are counted to reflect repeated visibility.
We then extract the overlapping domains for each \ac{aise} query along with their corresponding rank positions in the \acp{tse}.\footnote{Sponsored content and advertisements are excluded to focus solely on organic traffic.}
Finally, we calculate the frequency of each rank position across all search results for each \ac{aise}.
Given that user attention concentrates disproportionately on the top search results~\cite{seo_traffic}, this ranking position as a indicator of search-driven exposure.

\Cref{fig:domain_rank} illustrates the rank distribution of overlapping domains in the \acp{tse} results.
Surprisingly, we observe heavy skew in overlapping domains toward the top-1 \ac{tse} results. 
The sources \acp{aise} most frequently cite are those ranked first by traditional search engines 23.27\% in Bing and 14.53\% in Google. 
This suggests that for the overlapped search results, \acp{aise} tends to cite the domains that are also considered as the most important in \acp{tse}.

\pb{Traffic-Based Popularity.}
We then use the Tranco ranking~\cite{tranco} to assess the user popularity of the cited domains.
Tranco has been widely adopted in prior work~\cite{tranco_ref_1,tranco_ref_2,tranco_ref_3,trance_ref_4} and allows us to assess whether \acp{aise} prioritize high-traffic websites or rely on less popular sources.
Here, a lower rank indicates higher popularity.
\Cref{tab:tranco_baseline} reports the mean and median Tranco ranks for all domains cited by each search engine.
A Kruskal-Wallis test~\cite{mckight2010kruskal} is further conducted to confirm statistical differences 
($H = 58627.99$, $p < 0.001$). 
Post-hoc results show all \acp{aise} differ significantly from \acp{tse} ($p < 0.001$). 
Overall, except for ChatGPT, the \acp{aise} tend to return less popular domains (\ie higher rank) than the \acp{tse}. 
Notably, Copilot and Gemini exhibit substantially higher average Tranco ranks with over 22{,}000 above both \acp{tse}, which indicates a stronger inclination toward low-traffic domains. 
Such a tendency may negatively affect user trust, as search behavior research~\cite{pan2007google} shows that users typically perceive high-traffic domains as more authoritative. 
Beside, lower-traffic domains may also be less reliable, which motivates our investigation of domain credibility in \S\ref{sec:source_reliability}.

\begin{table}[t!]
\centering
\caption{Mean and median Tranco ranks across search engines, emphasizing the divergence $\Delta$ of \acp{aise} from the \ac{tse} baseline.}
\resizebox{0.95\columnwidth}{!}{
\begin{tabular}{llllll}
\hline
\textbf{Search Engine} & \textbf{Mean($\mu$)} & \textbf{Median} & 
$\Delta$ \textbf{vs Google ($\mu$)} & $\Delta$ \textbf{vs Bing ($\mu$)} \\
\hline
\textbf{Bing}        & 48124.95  & 2412.0  &  +670.16   &     0.00  \\
\textbf{Google}      & 41427.79  & 746.0   &     0.00   &  -697.16  \\
AIMode      & 43118.93  & 1809.0  &   +691.14  &  -1006.02 \\
ChatGPT     & 40442.48  & 713.0   & -985.31    & -1682.47  \\
Copilot     & 69054.10  & 3912.0  & +27626.31  & +20929.15 \\
Gemini      & 64121.03  & 3805.0  & +22693.24  & +15996.08 \\
Grok        & 42846.77  & 2507.0  &   +418.98  &  -1278.18 \\
Perplexity  & 48208.00  & 1362.0  &  +6780.21  &    +83.05 \\
\hline
\end{tabular}}
\label{tab:tranco_baseline}
\end{table}

\begin{small}
\summary{
1) \acp{aise} return fewer sources than \acp{tse} and share only 38\% of domains with \acp{tse}. Despite this smaller volume, \acp{aise} present a less concentrated domain distribution.
2) For overlapping domains, \acp{aise} prioritize those first-ranked by \acp{tse}. However, for news-relevant domains, \acp{aise} and \acp{tse} exposes distinct sources.
3) Overall, \acp{aise} (expect for ChatGPT) return domains with lower popularity than \acp{tse} from traffic perspectives, which may negatively affect user trust.
}
\end{small}

\section{Source Quality (RQ2)} \label{sec:rq2}

In the previous section, we showed that \acp{aise} return a more diverse set of domains than \acp{tse}. 
However, the shift in retrieval format alters the user’s role: instead of browsing multiple links, users receive an \ac{llm}-generated summary with only a few embedded references.
This places greater emphasis on the quality of those few sources.  
Accordingly, \textbf{RQ2} evaluates source quality through two axis: reliability and cyber threats.

\begin{figure*}[t]
    \centering
    \begin{minipage}{0.49\linewidth}
        \centering
        \includegraphics[width=\linewidth]{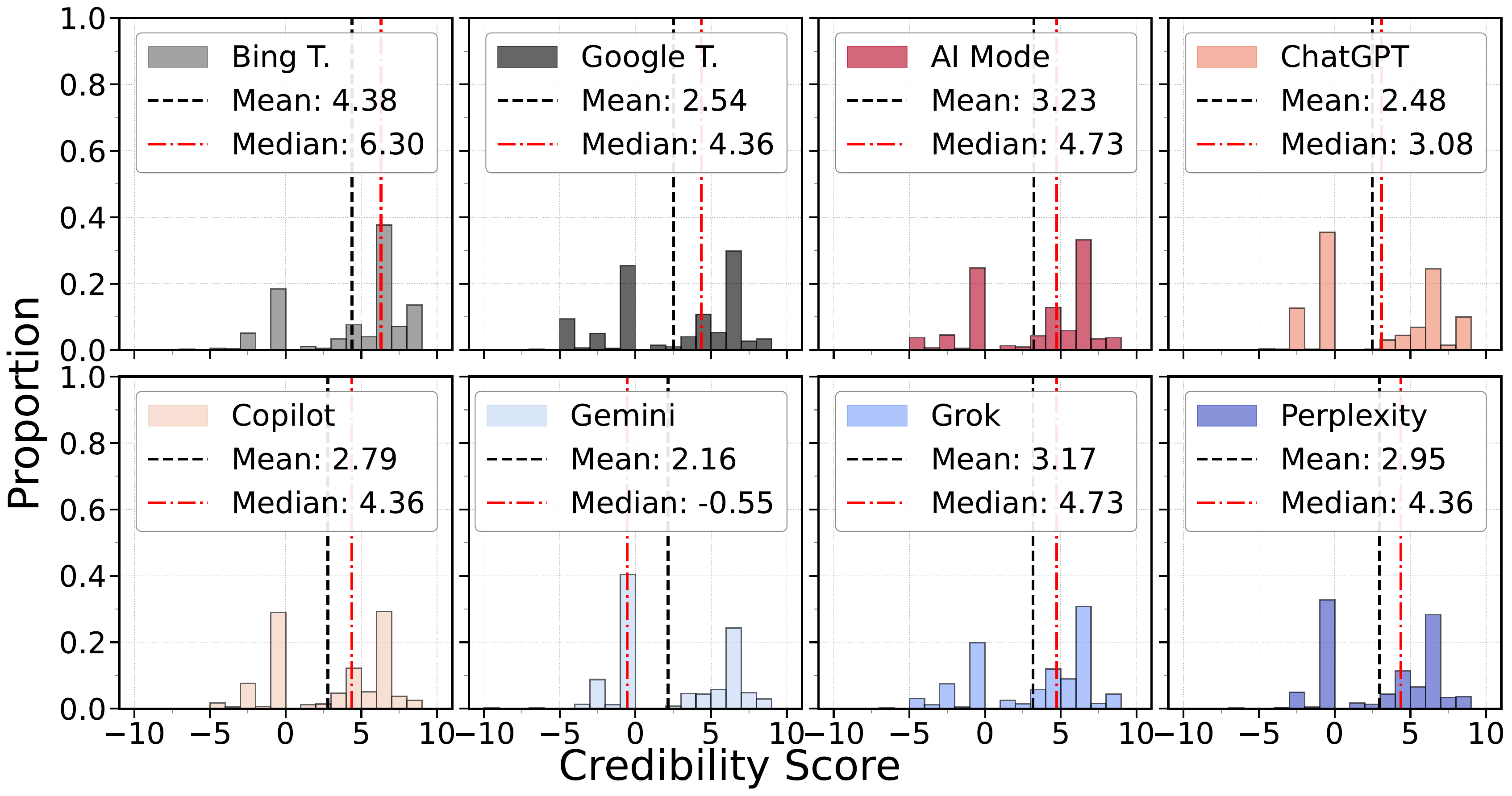}
        \subcaption{} 
        \label{fig:factual_score_histograms_final}
    \end{minipage}
         \begin{minipage}{0.49\linewidth}
        \centering
        \includegraphics[width=\linewidth]{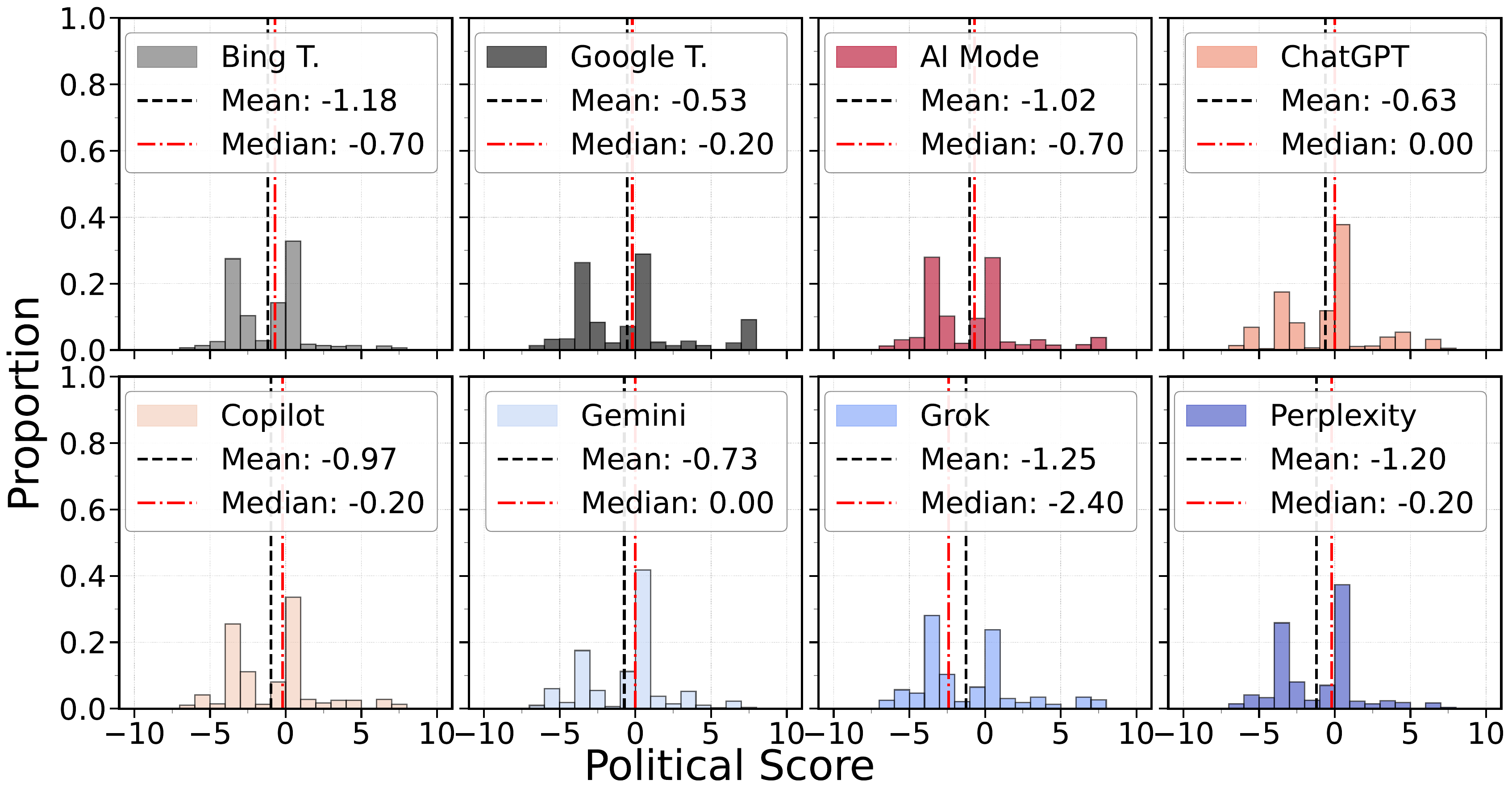}
        \subcaption{} 
        \label{fig:political_score_histograms_final}
    \end{minipage}
    \caption{Histograms of (a) credibility scores and (b) political scores for all domains based on \ac{mbfc}.}
    \label{fig:political_score}
\end{figure*}

\subsection{Reliability}
\label{sec:source_reliability}
We next study if \acp{aise} cite more reliable sources, as compared to \acp{tse}. 
We focus primarily on news and media landscapes, as these sources are most directly linked to information bias and shape users' perceptions of factual accuracy~\cite{human_trust_in_ai_search}.

\pb{Assigning Scores}
To evaluate this, we use the ratings from \ac{mbfc}~\cite{mdfc} to assess political leaning (\ie ownership and editor affiliation) and credibility (\ie transparency and accuracy) of cited domains.\footnote{Note, domains without \ac{mbfc} records are excluded. Detailed matching rates are shown in \S\ref{sec:data_collection_and_annotation}.}
\qty{51}{\percent} of ratings provide only categorical labels rather than the full scores.  
Following the official methodology~\cite{mdfc_method}, we therefore assign the average value of each category for the missing scores.  
Note that the original credibility scores range from $-1$ to $10$. 
To align with the political leaning scale ($-10$ to $10$) and facilitate visualization (with $0$ as the neutral midpoint), we transform credibility scores to range from $-10$ (least factual) to $10$ (most factual) (see Appendix \S\ref{app:mbfc_transform}).
Both the credibility and political scores are statistically significant according to the Kruskal--Wallis test (see \Cref{app:test_cp}). 
Such difference indicates that the distinct sourcing practices described in \S\ref{sec:disparity} indeed affect overall source quality.

\pb{Credibility.} 
We first examine the credibility of the response using the factual-reporting ratings from \ac{mbfc}~\cite{mdfc_method}.
\Cref{fig:factual_score_histograms_final} presents the credibility scores for all domains returned across search engines.
The prominent spike visible in all distributions is driven largely by Wikipedia (score: $-0.9$), which represents nearly one-third (\qty{32}{\percent}) of all domain occurrences. 
Bing outperform other search engines, primarily due to its pronounced concentrations around around \url{bing.com}  (score: $6.3$) and \url{cambridge.org} (score: $8.3$), comprising \qty{28}{\percent} and \qty{9}{\percent} of its cited domains, respectively.
As for \acp{aise}, except for Gemini and ChatGPT, other \acp{aise} fall between Bing and Google.  
However, such concentration of domains in response level also brings troubles.
For example, Gemini shows an exceptionally high reliance on Wikipedia (score: $-0.9$), which accounts for \qty{41}{\percent} of its citations, and \qty{62}{\percent} of its responses cite only a single \ac{mbfc}-listed domain. 
This narrow sourcing not only reduces variance but also corresponds to its lowest credibility scores (in both mean and median). 
These results suggest that greater source diversity is an important factor for improving credibility.

\pb{Political Leaning.} 
The political leaning of domains is rated by \acp{mbfc}~\cite{mdfc_method}.\footnote{Note, the score is termed ``Bias'' by \acp{mbfc}.}  
\Cref{fig:political_score_histograms_final}  illustrates that the political scores for all domains returned by the search engines range from -10 (left) to +10 (right).
Overall, given our set of queries, all engines display a left-leaning political orientation, with both mean and median scores below zero. 
Except for Grok, the mean is lower than the median for other search engines, which confirms a set of extreme left-leaning outliers.
This pattern reinforces the overall left-leaning tendency across engines.
For example, \qty{6.4}{\percent} of domains cited by Gemini fall within the far- or extreme-left range ($<-5$). 
This bias is exacerbated by Gemini's tendency to cite fewer sources than comparable systems (see \S\ref{sec:number_of_sources}), thereby concentrating user exposure on a narrow set of ideologically skewed domains. 
To capture the aggregate bias presented to users, we further evaluate political neutrality at the response level by computing the mean political score of all cited domains. 
Under this metric, Grok exhibits the lowest neutrality with only \qty{16}{\percent} of responses achieving a score of $0$, contradicting public claims of balanced positioning.\footnote{\url{https://www.nytimes.com/2025/09/02/technology/elon-musk-grok-conservative-chatbot.html}} 
Ultimately, the reliance of both Gemini and Grok on the fewest distinct domains underscores the critical role of source diversity in mitigating political bias.

\subsection{Cyber Threat}
\label{rq2_cyber_threats}
Finally, we assess the cyber threats to ensure that source credibility does not compromise user safety.  
For this, each domain is checked using VirusTotal,\footnote{\url{https://www.virustotal.com}} which aggregates reports from 97 security vendors~\cite{10.1145/3673660.3655042}.
To balance false positives and false negatives, we adopt a threshold of two or more positive flags to classify a domain as unsafe, following prior work~\cite{imc_virus_total_threshold}.
Alternative thresholds are evaluated in Appendix~\ref{app:virus_threshold}.  
This approach allows us to compare whether \acp{aise} are more likely than \acp{tse} to include unsafe sources.

\pb{Exposure Risk} 
Among 125{,}555 unique domains returned by search engines, 439 (\qty{0.36}{\percent}) are labeled as malicious. 
The distribution follows a log-scale dominated by a small subset of domains. 
The top \qty{10}{\percent} of malicious domains account for \qty{59}{\percent} of all occurrences, with the most frequent domain appearing 137 times.  
Within this dominant subset, Google (\qty{19}{\percent}) and Copilot (\qty{18}{\percent}) contributes the largest share of total malicious domains returned.  
Table~\ref{tab:cyber_threats} provides a further assessment of exposure risks (\ie the likelihood of encountering malicious domains) across search engines.
As baselines, Bing and Google \acp{tse} generate \qty{0.42}{\percent} and \qty{0.23}{\percent} malicious domains among all cited domains, respectively.  
Most \acp{aise} exhibit risks that fall between these two baselines, despite sourcing fewer number of domains per response.
We also assess the severity of maliciousness by measuring the average number of malicious domains per response.
Only Gemini show slightly higher averages compared to both of the baselines.  
Although Gemini and Google operate within the same ecosystem, Gemini yields a higher mean (Avg = 1.56, $\times$1.5 of Google).  
These findings suggest that while \acp{aise} may cite fewer sources overall, they do not necessarily mitigate exposure to risky domains.  
Moreover, engines within the same ecosystem can behave inconsistently in threat sourcing.  
This motivates us to further examine: 
\begin{enumerate*}
    \item whether overlapping sets of malicious domains appear across engines for the same query; and 
    \item which types of queries are more likely to trigger such risks.
\end{enumerate*}

\begin{table*}[ht]
    \centering
    \caption{Cyber Threat Exposure Risk Across Search Engines}
    \begin{tabular}{@{}lcccccccc@{}}
        \toprule
        \textbf{Metric} & \textbf{AI Mode} & \textbf{ChatGPT} & \textbf{Copilot} & \textbf{Gemini} & \textbf{Grok} & \textbf{Perplexity} & \textbf{Bing T.} & \textbf{Google T.} \\ \midrule
        
        \% of Threat Domains & \textbf{\qty{0.25}{\percent}} & \textbf{\qty{0.32}{\percent}} & \qty{0.34}{\percent} & \textbf{\qty{0.26}{\percent}} & \textbf{\qty{0.34}{\percent}} & \qty{0.23}{\percent} & \qty{0.42}{\percent} & \qty{0.23}{\percent} \\ 
        Avg. of Threat Domains per Resp. & \textbf{1.07} & 1.01 & 1.03 & \textit{\textbf{1.56}} & \textbf{1.11} & 1.02 & 1.33 & 1.05 \\
        \bottomrule
        \multicolumn{9}{@{}l}{\footnotesize Numbers that exceed either of the baselines from Google or Bing (\acp{tse}) are displayed in \textbf{bold}, while those exceeding both baselines are shown in \textbf{\textit{italic}}.  }
    \end{tabular}
    \label{tab:cyber_threats}
\end{table*}

\pb{Overlapping Domains.} 
We use the Jaccard Similarity to measure the similarity of domains between malicious responses generated for the same query across different search engines.
A Jaccard similarity above 0.5 is considered high and indicates substantial overlap between two sets~\cite{jaccard_support}.  
This analysis examines whether engines tend to return overlapping malicious sources.   
Overall, \qty{85}{\percent} of engine pairs share an average score above 0.5, and \qty{14}{\percent} exceed 0.8, indicating that the malicious domain sets are largely identical across engines for the same query.  
This high overlap in malicious domains implies that coordinated mitigation strategies targeting shared malicious sources could reduce the spread of harmful content across search engine optimization.

\pb{Search Query Category.}
We finally compute the Jaccard similarity for query sets that trigger malicious domains to assess whether specific types of queries are more prone to such threats.  
All Jaccard values fall below 0.4, indicating that malicious domains are triggered by distinct queries.  
However, potential overlaps in categories may still exist. 
To explore this, we employ an \ac{llm} to map each query to its corresponding category (details in \Cref{app:query_categorization}).  
Over \qty{50}{\percent} of queries that return malicious domains fall into four primary categories: \textit{arts \& entertainment}, \textit{people \& society}, \textit{finance}, and \textit{sports} (see \Cref{app:sus_query_cate}). 
This indicates that \acp{aise} may return malicious domains even for seemingly innocuous topics.

\begin{small}
\summary{
1) Gemini cites far-left domains in \qty{6.4}\% of cases, while Grok exhibits the lowest neutrality per response level among all \acp{aise}. 
2) All search engines return malicious domains even for seemingly innocuous queries. 
Despite returning fewer sources, \acp{aise} do not outperform \acp{tse}, and malicious domains often overlap across identical queries.
}
\end{small}
\section{\acp{aise} vs.\ \acp{tse} Website Characteristics}
\label{sec:rq3}
From the previous analysis, we confirm that \acp{aise} represents a new search paradigm that diverges from \acp{tse} ranking mechanisms. 
To better understand the mechanisms underlying \acp{aise} citation behavior, we conduct a feature-based analysis to identify key factors that influence source selection in \acp{aise}.

\subsection{Features in HTML}
We begin by analyzing key \ac{html}-based features.  
Prior studies~\cite{tan2025htmlrag,lastowka2000search,thurow2003search} show that HTML structure strongly influences how both \acp{aise} and \acp{tse} retrieve and cite content.  
This influence arises from two aspects: the plain text and the underlying structure of HTML~\cite{tan2025htmlrag}. Accordingly, we analyze both types of HTML features. 

We randomly sample 10{,}000 URLs from each comparison set.\footnote{Note that all selected domains explicitly permit data crawling, as verified via their \texttt{robots.txt}.}
For the plain text, we assess the readability of the content and compute two widely used readability metrics: Flesch Reading Ease Score~\cite{kincaid1975derivation} and Flesch–Kincaid Grade Level~\cite{flesch2007flesch}.
For the \ac{html} structure, we extract four features: semantic tags, DOM nesting depth, accessibility attributes, and counts of deprecated or discouraged tags.
These metrics are detailed in~\Cref{app:html_structure}.

\begin{table}[ht]
\centering
\caption{Kolmogorov-Smirnov (KS) Test Results for Feature Distributions Between Groups.}
\label{tab:ks_results}
\resizebox{0.9\columnwidth}{!}{
\begin{tabular}{llll}
\toprule
\textbf{Feature} & \textbf{KS} & \textbf{p-value} & \textbf{Mean}\\
\midrule
Semantic Tags           & 0.061 & \textbf{$p < 0.01$} & \acp{aise}>\acp{tse} \\
Nesting Depth           & 0.069 & \textbf{$p < 0.01$} & \acp{aise}>\acp{tse} \\
Accessibility Features  & 0.066 & \textbf{$p < 0.01$} & \acp{aise}=\acp{tse} \\
Markup Errors           & 0.022 & \textbf{$p < 0.01$} & \acp{aise}=\acp{tse} \\
Flesch Reading Ease Score  &  0.096 & \textbf{$p < 0.01$} & \acp{aise}>\acp{tse} \\
Flesch–Kincaid Grade Level  & 0.082 & \textbf{$p < 0.01$} & \acp{aise}<\acp{tse}  \\
\bottomrule
\end{tabular}}
\end{table}

To assess the distributional differences between \acp{aise} and \acp{tse}, we use the Kolmogorov–Smirnov (KS) test to compare their empirical distributions.
\Cref{tab:ks_results} presents the KS statistics and mean comparisons for the six features.
All differences are statistically significant ($p<0.01$) across both textual readability and structural characteristics.

In terms of readability, \acp{tse} content is considerably more complex, with an average grade level of 18.24 (vs.\ 14.57 for \acp{aise}) and a reading ease score of 12.32 (vs.\ 24.15).  
These results suggest that \acp{aise} tend to favor more accessible, less textually demanding content when selecting sources.
For structural features, \acp{aise} pages show a modest increase in HTML tag usage (mean: 7 vs.\ 6) and nesting depth (mean: 16 vs.\ 15) compared to \acp{tse} pages.
The nesting depth feature measures the maximum level of HTML tag embedding within a webpage, reflecting the structural complexity of its DOM hierarchy.
Overall, \acp{aise} appear to prioritize sources that are both more readable and structurally informative, reflecting a preference for content that is easier for both humans and machines to process.

\subsection{Predicting Unique Inclusion of Domains}
Beyond HTML features, our analyses in \S\ref{sec:rq2} indicate that domain-level annotation features may also influence sourcing preferences.  
To quantify their impact, we train predictive models and examine feature importance, identifying the factors that most strongly differentiate domains that are uniquely cited by \acp{aise} from those cited by \acp{tse}.

\begin{table}[]
\caption{Overview of input features and preprocessing steps used in the classification pipeline.}
\label{tab:features}
\resizebox{0.9\columnwidth}{!}{
\begin{tabular}{@{}lll@{}}
\toprule
\textbf{Feature Type} & \textbf{Feature Name(s)} & \textbf{Transformation}             \\ 
\midrule
Numeric               & \begin{tabular}[c]{@{}l@{}}Tranco ranking,\\ the number of out-links from VT\\MBFC bias and factual
\end{tabular} & \begin{tabular}[c]{@{}l@{}}Mean imputation \\ Standard Scaler\end{tabular}            \\
Categorical           & domain, media category from VT                                                         & \begin{tabular}[c]{@{}l@{}}Most frequent imputation  \\ One-Hot Encoding\end{tabular} \\
Textual                & search query, description from VT                                                                 & \begin{tabular}[c]{@{}l@{}}TF-IDF\end{tabular} \\ \bottomrule
\end{tabular}}
\end{table}

\pb{Classifier Design.}
We train a classifier to differentiate domains that uniquely appear in \acp{aise} from those that are exclusive to \acp{tse}, aiming to identify the characteristics that distinguish their respective source selections.
The input features are structured into three modalities: numeric, categorical, and textual shows in Table~\ref{tab:features}. 
We perform a stratified 80:20 train–test split, and select models using 5-fold stratified cross-validation on the training set.  
Five classifier types are evaluated, with hyperparameter tuning conducted via grid search to optimize performance.  
Detailed hyperparameter configurations are provided in \Cref{app:hyper}.

\pb{Model Evaluation.}
Table~\ref{tab:f1_detailed} reports per-class F1 scores for cited and non-cited labels, along with the overall weighted F1 score.  
The best-performing model is XGBoost, achieving a weighted F1 score of 0.758. The model performs notably better in identifying domains unique to \acp{aise}, with a precision of 0.668 and a high recall of 0.879, suggesting strong sensitivity in detecting such domains.

\begin{table}[]
\caption{Per-class and overall weighted F1 scores for predicting whether a domain is cited exclusively by \acp{aise}.}
\resizebox{.7\columnwidth}{!}{
\begin{tabular}{@{}lllllll@{}}
\toprule
                     & \textbf{LR} & \textbf{RF} & \textbf{XGBoost} & \textbf{KNN} & \textbf{MLP} \\ \midrule
\textbf{AI-SE}       & 0.693       & 0.740       & 0.759            & 0.697     &  0.708  \\
\textbf{T-SE}        & 0.576       & 0.505       & 0.450            & 0.439     &  0.519   \\
\textbf{F1}          & 0.692       & 0.727       & 0.758            & 0.697     &  0.707   \\ \bottomrule
\end{tabular}}
\label{tab:f1_detailed}
\end{table}

\pb{Feature Importance.}
Finally, we analyze the feature importance of the best-performing model, XGBoost, using the SHAP framework~\cite{shap}. 
Six features have mean SHAP values exceeding 0.1. The most influential are the Tranco ranking (0.923), number of outlinks (0.799), and the .com top-level domain (0.623), followed by subdomain count (0.189), .nl domains (0.135), and the business and economy category (0.116). 
These features collectively exert a strong positive influence on the model output.
The Tranco ranking captures a domain’s global popularity based on aggregated web traffic rankings. 
Note, a higher numerical rank corresponds to lower popularity comparing.
This aligns with our findings in Section~\ref{sec:rq2} \acp{aise} favor for less popular domains comparing with \acp{tse}.
The number of outlinks represents how many external references a webpage contains; a larger number of outlinks may reflect richer contextualization and stronger integration with other resources.

\begin{small}
\summary{
Domains favored by \acp{aise} generally exhibit four key characteristics: 1) structured HTML with deeper and more hierarchical layouts, 2) more readable textual content, 3) higher Tranco ranking (less popular websites), and 4) more outlinks.
}
\end{small}

\section{Conclusion and Future work}

\pb{Conclusion.}
We present the first large-scale empirical study of the sourcing behaviors of \acp{aise}, covering 55{,}936 question across six \acp{aise} and two \acp{tse}. 
\S\ref{sec:rq1} reveals that \acp{aise} diverge from the sourcing paradigm of \acp{tse}.
\acp{aise} return far fewer URLs per response ($4.3$ vs.\ $10$ on average), yet \qty{37}{\percent} of the domains are absent from \acp{tse}.
\S\ref{sec:rq2} further assesses the sourcing quality of this novel search behavior across two dimensions: \emph{reliability} and \emph{cyber threats}.
We find that \acp{aise} return less popular domains, on average.
Thus, overall reliability depends heavily on these selections, with source diversity serving as a key indicator of credibility.
Moreover, despite returning far fewer sources, \acp{aise} exhibit risk levels comparable to those of \acp{tse}.
Finally, \S\ref{sec:rq3} identifies the features influencing source selections: \acp{aise} are biased toward established, well-referenced information environments.
This potentially improving citation quality yet narrowing the diversity of accessible content.
Together, these findings highlight the need for more diverse and safer sourcing practices in future \acp{aise}.

\pb{Future Work.}
The architecture of \acp{aise} is multifaceted, encompassing document retrieval, internal ranking, summarization, and citation selection. 
Our analysis restricts its scope to user-facing citations to best assess the impact on end-users, yet lacking the measuring the intermediate retrieval pipeline effectively.
Concurrently, commercial imperatives (\eg sponsored content) raise concerns regarding the influence of financial incentives on information prioritization. 
We identify two critical avenues for future research. 
First, future work could audit full retrieval traces to reveal potential biases in pre-generation filtering and ranking. 
For example, we notice that some \acp{aise} (\eg Grok) disclose their internal reasoning and retrieval steps recently, which offer us a new opportunity to examine the factors underlying their pre-generation process.
Second, research could examine monetization-driven behaviors, such as Generative Engine Optimization (GEO), to understand how commercial incentives influence content delivery. 
Notably, ChatGPT launched a shopping assistant in their recent product.\footnote{\url{https://openai.com/index/chatgpt-shopping-research/}} 
GEO providers can boost the visibility of certain brands in the outputs by hijacking specific keywords, thereby influencing the final source selection.
Exploring these directions will clarify the impact of \acp{aise} on transparency and information integrity.

\clearpage
\bibliographystyle{acm}
\bibliography{ref}

@String{Computing = "Computing" }

@String{Computer = "{IEEE} Computer" }

@String{Springer = "Springer-Verlag" }

@misc{tf_idf,
  author       = {Wikipedia},
  title        = {Tf-idf Algorithms},
  howpublished = {\url{https://en.wikipedia.org/wiki/Tf%E2%80%93idf}}
}

@misc{genai_survey,
  author       = {Statista},
  title        = {Generative AI Search Survey},
  howpublished = {\url{https://www.statista.com/statistics/1454204/united-states-generative-ai-primary-usage-online-search/}}
}

@misc{hits,
  author       = {Wikipedia},
  title        = {HITS Algorithms},
  howpublished = {\url{https://en.wikipedia.org/wiki/HITS_algorithm}},
}

@misc{pagerank,
  author       = {Wikipedia},
  title        = {Page rank Algorithms},
  howpublished = {\url{https://en.wikipedia.org/wiki/PageRank}},
}

@misc{sitemap,
  author       = {Google Inc.},
  title        = {Site Map},
  howpublished ={\url{https://developers.google.com/search/docs/crawling-indexing/sitemaps/overview}},
}

@misc{virustotal,
  author       = {Hispasec Sistemas},
  title        = {VirusTotal Documentation},
  howpublished ={\url{https://docs.virustotal.com/docs/results-reports}},
}

@misc{google_search,
  author       = {Google Inc.},
  title        = {Google Search Development Document},
  howpublished ={\url{https://developers.google.com/search}},
}

@misc{bing_search,
  author       = {Microsoft Corporation},
  title        = {Bing Search Development Document},
  howpublished ={\url{https://www.bing.com/webmasters/help/webmaster-guidelines-30fba23a}},
}

@misc{mdfc,
    author       = {Dave M. Van Zandt},
    title        = {Media Bias Fact Check Official Website},
    howpublished = {\url{https://mediabiasfactcheck.com}},
}

@inproceedings{imc_virus_total_threshold,
author = {Peng, Peng and Yang, Limin and Song, Linhai and Wang, Gang},
title = {Opening the Blackbox of VirusTotal: Analyzing Online Phishing Scan Engines},
year = {2019},
isbn = {9781450369480},
publisher = {Association for Computing Machinery},
address = {New York, NY, USA},
url = {https://doi.org/10.1145/3355369.3355585},
doi = {10.1145/3355369.3355585},
booktitle = {Proceedings of the Internet Measurement Conference},
pages = {478–485},
numpages = {8},
location = {Amsterdam, Netherlands},
series = {IMC '19}
}

@article{10.1145/3673660.3655042,
author = {Choo, Euijin and Nabeel, Mohamed and Kim, Doowon and De Silva, Ravindu and Yu, Ting and Khalil, Issa},
title = {A Large Scale Study and Classification of VirusTotal Reports on Phishing and Malware URLs},
year = {2024},
issue_date = {June 2024},
publisher = {Association for Computing Machinery},
address = {New York, NY, USA},
volume = {52},
number = {1},
issn = {0163-5999},
url = {https://doi.org/10.1145/3673660.3655042},
doi = {10.1145/3673660.3655042},
journal = {SIGMETRICS Perform. Eval. Rev.},
month = jun,
pages = {55–56},
numpages = {2}
}

@article{tranco_ref_1,
  title={A survey on malware detection with graph representation learning},
  author={Bilot, Tristan and El Madhoun, Nour and Al Agha, Khaldoun and Zouaoui, Anis},
  journal={ACM Computing Surveys},
  volume={56},
  number={11},
  pages={1--36},
  year={2024},
  publisher={ACM New York, NY}
}

@inproceedings{tranco_ref_2,
  title={Design and evaluation of IPFS: a storage layer for the decentralized web},
  author={Trautwein, Dennis and Raman, Aravindh and Tyson, Gareth and Castro, Ignacio and Scott, Will and Schubotz, Moritz and Gipp, Bela and Psaras, Yiannis},
  booktitle={Proceedings of the ACM SIGCOMM 2022 Conference},
  pages={739--752},
  year={2022}
}

@inproceedings{tranco_ref_3,
  title={A browser-side view of starlink connectivity},
  author={Kassem, Mohamed M and Raman, Aravindh and Perino, Diego and Sastry, Nishanth},
  booktitle={Proceedings of the 22nd ACM Internet Measurement Conference},
  pages={151--158},
  year={2022}
}

@article{trance_ref_4,
  title={A comparative study of dark patterns across web and mobile modalities},
  author={Gunawan, Johanna and Pradeep, Amogh and Choffnes, David and Hartzog, Woodrow and Wilson, Christo},
  journal={Proceedings of the ACM on Human-Computer Interaction},
  volume={5},
  number={CSCW2},
  pages={1--29},
  year={2021},
  publisher={ACM New York, NY, USA}
}

@inproceedings{bouwman2022helping,
  title={Helping hands: Measuring the impact of a large threat intelligence sharing community},
  author={Bouwman, Xander and Le Pochat, Victor and Foremski, Pawel and Van Goethem, Tom and Ga{\~n}{\'a}n, Carlos H and Moura, Giovane CM and Tajalizadehkhoob, Samaneh and Joosen, Wouter and Van Eeten, Michel},
  booktitle={31st USENIX Security Symposium (USENIX Security 22)},
  pages={1149--1165},
  year={2022}
}

@inproceedings{10.5555/3359012.3359022,
author = {Le Pochat, Victor and Van Goethem, Tom and Joosen, Wouter},
title = {Evaluating the long-term effects of parameters on the characteristics of the Tranco top sites ranking},
year = {2019},
publisher = {USENIX Association},
address = {USA},
booktitle = {Proceedings of the 12th USENIX Conference on Cyber Security Experimentation and Test},
pages = {10},
numpages = {1},
location = {Santa Clara, CA, USA},
series = {CSET'19}
}

@inproceedings{10.1145/3711542.3711601,
author = {Shah, Bhushan Santosh and Shah, Deven Santosh and Attar, Vahida},
title = {Decoding News Bias: Multi Bias Detection in News Articles},
year = {2025},
isbn = {9798400717383},
publisher = {Association for Computing Machinery},
address = {New York, NY, USA},
url = {https://doi.org/10.1145/3711542.3711601},
doi = {10.1145/3711542.3711601},
booktitle = {Proceedings of the 2024 8th International Conference on Natural Language Processing and Information Retrieval},
pages = {97–104},
numpages = {8},
series = {NLPIR '24}
}

@misc{miroyan2025searcharenaanalyzingsearchaugmented,
      title={Search Arena: Analyzing Search-Augmented LLMs}, 
      author={Mihran Miroyan and Tsung-Han Wu and Logan King and Tianle Li and Jiayi Pan and Xinyan Hu and Wei-Lin Chiang and Anastasios N. Angelopoulos and Trevor Darrell and Narges Norouzi and Joseph E. Gonzalez},
      year={2025},
      eprint={2506.05334},
      archivePrefix={arXiv},
      primaryClass={cs.CL},
      url={https://arxiv.org/abs/2506.05334}, 
}

@book{thurow2003search,
  title={Search engine visibility},
  author={Thurow, Shari},
  year={2003},
  publisher={New Riders}
}

@article{lastowka2000search,
  title={Search engines, HTML, and trademarks: What's the meta for},
  author={Lastowka, F Gregory},
  journal={Va. L. Rev.},
  volume={86},
  pages={835},
  year={2000},
  publisher={HeinOnline}
}

@misc{forcepoint,
    author       = {Forcepoint ThreatSeeker},
    title        = {Forcepoint ThreatSeeker Document},
    howpublished = {\url{https://help.forcepoint.com/fpone/migration/webtothreatseekerurl/guid-b8ac3928-a64c-4468-bf16-4a6d4932c2c7.html/}}
}

@misc{drissionPage,
    author       = {DrissionPage},
    title        = {DrissionPage Repository},
    howpublished = {\url{https://github.com/g1879/DrissionPage}}
}

@techreport{2012-dittrich-mraf,
  author = {Dittrich, D and Kenneally, E},
  title = {{The Menlo Report: Ethical Principles Guiding Information and Communication Technology Research}},
  institution = {U.S. Department of Homeland Security},
  year = {2012},
  month = {August},
  doi = {https://catalog.caida.org/paper/2012_menlo_report_actual_formatted}
}

@misc{google_trend,
    author       = {Google Inc.},
    title        = {Google Trend Document},
    howpublished = {\url{https://developers.google.com/search/docs/monitor-debug/trends-start}}
}

@misc{human_trust_in_ai_search,
      title={Human Trust in AI Search: A Large-Scale Experiment}, 
      author={Haiwen Li and Sinan Aral},
      year={2025},
      eprint={2504.06435},
      archivePrefix={arXiv},
      primaryClass={cs.CY},
      url={https://arxiv.org/abs/2504.06435}, 
}

@article{gastwirth1971general,
  title={A general definition of the Lorenz curve},
  author={Gastwirth, Joseph L},
  journal={Econometrica: Journal of the Econometric Society},
  pages={1037--1039},
  year={1971},
  publisher={JSTOR}
}

@inproceedings{10.1145/3726302.3730230,
author = {Avula, Sandeep and Lee, Chia-Jung and Zhang, Rongting and Murdock, Vanessa},
title = {Measuring the Fairness Gap Between Retrieval and Generation in RAG Systems using a Cognitive Complexity Framework},
year = {2025},
isbn = {9798400715921},
publisher = {Association for Computing Machinery},
address = {New York, NY, USA},
url = {https://doi.org/10.1145/3726302.3730230},
doi = {10.1145/3726302.3730230},
booktitle = {Proceedings of the 48th International ACM SIGIR Conference on Research and Development in Information Retrieval},
pages = {2994–2998},
numpages = {5},
location = {Padua, Italy},
series = {SIGIR '25}
}

@inproceedings{giorgi2025human,
  title={Human and LLM biases in hate speech annotations: A socio-demographic analysis of annotators and targets},
  author={Giorgi, Tommaso and Cima, Lorenzo and Fagni, Tiziano and Avvenuti, Marco and Cresci, Stefano},
  booktitle={Proceedings of the International AAAI Conference on Web and Social Media},
  volume={19},
  pages={653--670},
  year={2025}
}

@article{huang2025survey,
  title={A survey on hallucination in large language models: Principles, taxonomy, challenges, and open questions},
  author={Huang, Lei and Yu, Weijiang and Ma, Weitao and Zhong, Weihong and Feng, Zhangyin and Wang, Haotian and Chen, Qianglong and Peng, Weihua and Feng, Xiaocheng and Qin, Bing and others},
  journal={ACM Transactions on Information Systems},
  volume={43},
  number={2},
  pages={1--55},
  year={2025},
  publisher={ACM New York, NY}
}

@article{10.1145/3744746,
author = {Naveed, Humza and Khan, Asad Ullah and Qiu, Shi and Saqib, Muhammad and Anwar, Saeed and Usman, Muhammad and Akhtar, Naveed and Barnes, Nick and Mian, Ajmal},
title = {A Comprehensive Overview of Large Language Models},
year = {2025},
issue_date = {October 2025},
publisher = {Association for Computing Machinery},
address = {New York, NY, USA},
volume = {16},
number = {5},
issn = {2157-6904},
url = {https://doi.org/10.1145/3744746},
doi = {10.1145/3744746},
journal = {ACM Trans. Intell. Syst. Technol.},
month = aug,
articleno = {106},
numpages = {72}
}

@misc{eeat,
    author       = {Search Engine Journal},
    title        = {Google EEAT: What Is It \& How To Demonstrate It For SEO},
    howpublished = {\url{https://www.searchenginejournal.com/google-e-e-a-t-how-to-demonstrate-first-hand-experience/474446/}}
}

@inproceedings{yao2022react,
  title={React: Synergizing reasoning and acting in language models},
  author={Yao, Shunyu and Zhao, Jeffrey and Yu, Dian and Du, Nan and Shafran, Izhak and Narasimhan, Karthik R and Cao, Yuan},
  booktitle={The eleventh international conference on learning representations},
  year={2022}
}

@article{lewis2020retrieval,
  title={Retrieval-augmented generation for knowledge-intensive nlp tasks},
  author={Lewis, Patrick and Perez, Ethan and Piktus, Aleksandra and Petroni, Fabio and Karpukhin, Vladimir and Goyal, Naman and K{\"u}ttler, Heinrich and Lewis, Mike and Yih, Wen-tau and Rockt{\"a}schel, Tim and others},
  journal={Advances in neural information processing systems},
  volume={33},
  pages={9459--9474},
  year={2020}
}

@article{kabir2023answers,
  title={Who answers it better? an in-depth analysis of chatgpt and stack overflow answers to software engineering questions},
  author={Kabir, Samia and Udo-Imeh, David N and Kou, Bonan and Zhang, Tianyi},
  journal={CoRR},
  year={2023}
}

@inproceedings{press2023measuring,
  title={Measuring and Narrowing the Compositionality Gap in Language Models},
  author={Press, Ofir and Zhang, Muru and Min, Sewon and Schmidt, Ludwig and Smith, Noah A. and Lewis, Mike},
  booktitle={Findings of the Association for Computational Linguistics: EMNLP 2023},
  pages={5687--5711},
  year={2023},
  publisher={Association for Computational Linguistics}
}

@inproceedings{gao-etal-2023-enabling,
    title = "Enabling Large Language Models to Generate Text with Citations",
    author = "Gao, Tianyu  and
      Yen, Howard  and
      Yu, Jiatong  and
      Chen, Danqi",
    editor = "Bouamor, Houda  and
      Pino, Juan  and
      Bali, Kalika",
    booktitle = "Proceedings of the 2023 Conference on Empirical Methods in Natural Language Processing",
    month = dec,
    year = "2023",
    address = "Singapore",
    publisher = "Association for Computational Linguistics",
    url = "https://aclanthology.org/2023.emnlp-main.398/",
    doi = "10.18653/v1/2023.emnlp-main.398",
    pages = "6465--6488"
}

@article{nakano2021webgpt,
  title={Webgpt: Browser-assisted question-answering with human feedback},
  author={Nakano, Reiichiro and Hilton, Jacob and Balaji, Suchir and Wu, Jeff and Ouyang, Long and Kim, Christina and Hesse, Christopher and Jain, Shantanu and Kosaraju, Vineet and Saunders, William and others},
  journal={arXiv preprint arXiv:2112.09332},
  year={2021}
}

@article{menick2022teaching,
  title={Teaching language models to support answers with verified quotes},
  author={Menick, Jacob and Trebacz, Maja and Mikulik, Vladimir and Aslanides, John and Song, Francis and Chadwick, Martin and Glaese, Mia and Young, Susannah and Campbell-Gillingham, Lucy and Irving, Geoffrey and others},
  journal={arXiv preprint arXiv:2203.11147},
  year={2022}
}

@inproceedings{trivedi2023interleaving,
  title={Interleaving retrieval with chain-of-thought reasoning for knowledge-intensive multi-step questions},
  author={Trivedi, Harsh and Balasubramanian, Niranjan and Khot, Tushar and Sabharwal, Ashish},
  booktitle={Proceedings of the 61st annual meeting of the association for computational linguistics (volume 1: long papers)},
  pages={10014--10037},
  year={2023}
}

@article{zhou2022least,
  title={Least-to-most prompting enables complex reasoning in large language models},
  author={Zhou, Denny and Sch{\"a}rli, Nathanael and Hou, Le and Wei, Jason and Scales, Nathan and Wang, Xuezhi and Schuurmans, Dale and Cui, Claire and Bousquet, Olivier and Le, Quoc and others},
  journal={arXiv preprint arXiv:2205.10625},
  year={2022}
}

@inproceedings{guan2007eye,
  title={An eye tracking study of the effect of target rank on web search},
  author={Guan, Zhiwei and Cutrell, Edward},
  booktitle={Proceedings of the SIGCHI conference on Human factors in computing systems},
  pages={417--420},
  year={2007}
}

@article{pan2007google,
  title={In Google we trust: Users’ decisions on rank, position, and relevance},
  author={Pan, Bing and Hembrooke, Helene and Joachims, Thorsten and Lorigo, Lori and Gay, Geri and Granka, Laura},
  journal={Journal of computer-mediated communication},
  volume={12},
  number={3},
  pages={801--823},
  year={2007},
  publisher={Oxford University Press Oxford, UK}
}

@misc{seo_traffic,
    author       = {SEO Inc.},
    title        = {How Much Traffic Is From Organic Search: (Updated for 2025)},
    howpublished ={\url{https://www.seoinc.com/seo-blog/much-traffic-comes-organic-search/}}
}

@misc{mdfc_method,
    author       = {Dave M. Van Zandt},
    title        = {Media Bias Fact Check Methodology},
    howpublished = {\url{https://mediabiasfactcheck.com/methodology/}}
}

@article{tranco,
  author       = {Victor Le Pochat and
                  Tom van Goethem and
                  Wouter Joosen},
  title        = {Rigging Research Results by Manipulating Top Websites Rankings},
  journal      = {CoRR},
  volume       = {abs/1806.01156},
  year         = {2018},
  url          = {http://arxiv.org/abs/1806.01156},
  eprinttype    = {arXiv},
  eprint       = {1806.01156},
  bibsource    = {dblp computer science bibliography, https://dblp.org}
}

@misc{lewis2021retrievalaugmentedgenerationknowledgeintensivenlp,
      title={Retrieval-Augmented Generation for Knowledge-Intensive NLP Tasks}, 
      author={Patrick Lewis and Ethan Perez and Aleksandra Piktus and Fabio Petroni and Vladimir Karpukhin and Naman Goyal and Heinrich Küttler and Mike Lewis and Wen-tau Yih and Tim Rocktäschel and Sebastian Riedel and Douwe Kiela},
      year={2021},
      eprint={2005.11401},
      archivePrefix={arXiv},
      primaryClass={cs.CL},
      url={https://arxiv.org/abs/2005.11401}, 
}

@inproceedings{hannak2013measuring,
  title={Measuring personalization of web search},
  author={Hannak, Aniko and Sapiezynski, Piotr and Molavi Kakhki, Arash and Krishnamurthy, Balachander and Lazer, David and Mislove, Alan and Wilson, Christo},
  booktitle={Proceedings of the 22nd international conference on World Wide Web},
  pages={527--538},
  year={2013}
}

@inproceedings{poudel2024navigating,
  title={Navigating the post-api dilemma},
  author={Poudel, Amrit and Weninger, Tim},
  booktitle={Proceedings of the ACM Web Conference 2024},
  pages={2476--2484},
  year={2024}
}

@article{dodds2023allotaxonometry,
  title={Allotaxonometry and rank-turbulence divergence: A universal instrument for comparing complex systems},
  author={Dodds, Peter Sheridan and Minot, Joshua R and Arnold, Michael V and Alshaabi, Thayer and Adams, Jane Lydia and Dewhurst, David Rushing and Gray, Tyler J and Frank, Morgan R and Reagan, Andrew J and Danforth, Christopher M},
  journal={EPJ Data Science},
  volume={12},
  number={1},
  pages={37},
  year={2023},
  publisher={Springer Berlin Heidelberg}
}

@inproceedings{tranco_reputation,
   author = "{Le Pochat}, Victor and {Van Goethem}, Tom and Tajalizadehkhoob, Samaneh and Korczy\'{n}ski, Maciej and Joosen, Wouter",
    title = "Tranco: A Research-Oriented Top Sites Ranking Hardened Against Manipulation",
booktitle = {Proceedings of the 26th Annual Network and Distributed System Security Symposium},
   series = {NDSS 2019},
     year = {2019},
    month = feb,
      doi = {10.14722/ndss.2019.23386},
}

@inproceedings{ieong2012domain,
  title={Domain bias in web search},
  author={Ieong, Samuel and Mishra, Nina and Sadikov, Eldar and Zhang, Li},
  booktitle={Proceedings of the fifth ACM international conference on Web search and data mining},
  pages={413--422},
  year={2012}
}

@inproceedings{sun2023delphi,
  title={DELPHI: Data for Evaluating LLMs’ Performance in Handling Controversial Issues},
  author={Sun, David and Abzaliev, Artem and Kotek, Hadas and Klein, Christopher and Xiu, Zidi and Williams, Jason D},
  booktitle={Proceedings of the 2023 Conference on Empirical Methods in Natural Language Processing: Industry Track},
  pages={820--827},
  year={2023}
}

@article{luo2025unsafe,
  title={Unsafe LLM-Based Search: Quantitative Analysis and Mitigation of Safety Risks in AI Web Search},
  author={Luo, Zeren and Peng, Zifan and Liu, Yule and Sun, Zhen and Li, Mingchen and Zheng, Jingyi and He, Xinlei},
  journal={arXiv preprint arXiv:2502.04951},
  year={2025}
}

@article{caramancion2024large,
  title={Large language models vs. search engines: evaluating user preferences across varied information retrieval scenarios},
  author={Caramancion, Kevin Matthe},
  journal={arXiv preprint arXiv:2401.05761},
  year={2024}
}

@inproceedings{vallina2020mis,
  title={Mis-shapes, mistakes, misfits: An analysis of domain classification services},
  author={Vallina, Pelayo and Le Pochat, Victor and Feal, {\'A}lvaro and Paraschiv, Marius and Gamba, Julien and Burke, Tim and Hohlfeld, Oliver and Tapiador, Juan and Vallina-Rodriguez, Narseo},
  booktitle={Proceedings of the ACM Internet Measurement Conference},
  pages={598--618},
  year={2020}
}

@book{jaccard_support,
  title={Mining of massive datasets},
  author={Rajaraman, Anand and Ullman, Jeffrey D},
  year={2011},
  publisher={Autoedicion}
}

@incollection{nielsen2016news,
  title={News media, search engines and social networking sites as varieties of online gatekeepers},
  author={Nielsen, Rasmus Kleis},
  booktitle={Rethinking journalism again},
  pages={93--108},
  year={2016},
  publisher={Routledge}
}

@article{haim2018burst,
  title={Burst of the filter bubble? Effects of personalization on the diversity of Google News},
  author={Haim, Mario and Graefe, Andreas and Brosius, Hans-Bernd},
  journal={Digital journalism},
  volume={6},
  number={3},
  pages={330--343},
  year={2018},
  publisher={Taylor \& Francis}
}

@inproceedings{spatharioti2025effects,
  title={Effects of LLM-based Search on Decision Making: Speed, Accuracy, and Overreliance},
  author={Spatharioti, Sofia Eleni and Rothschild, David and Goldstein, Daniel G and Hofman, Jake M},
  booktitle={Proceedings of the 2025 CHI Conference on Human Factors in Computing Systems},
  pages={1--15},
  year={2025}
}

@article{gezici2021evaluation,
  title={Evaluation metrics for measuring bias in search engine results},
  author={Gezici, Gizem and Lipani, Aldo and Saygin, Yucel and Yilmaz, Emine},
  journal={Information Retrieval Journal},
  volume={24},
  number={2},
  pages={85--113},
  year={2021},
  publisher={Springer}
}

@inproceedings{trielli2019search,
  title={Search as news curator: The role of Google in shaping attention to news information},
  author={Trielli, Daniel and Diakopoulos, Nicholas},
  booktitle={Proceedings of the 2019 CHI Conference on human factors in computing systems},
  pages={1--15},
  year={2019}
}

@inproceedings{wazzan2024comparing,
  title={Comparing traditional and LLM-based search for image geolocation},
  author={Wazzan, Albatool and MacNeil, Stephen and Souvenir, Richard},
  booktitle={Proceedings of the 2024 Conference on Human Information Interaction and Retrieval},
  pages={291--302},
  year={2024}
}

@inproceedings{dai2024bias,
  title={Bias and unfairness in information retrieval systems: New challenges in the llm era},
  author={Dai, Sunhao and Xu, Chen and Xu, Shicheng and Pang, Liang and Dong, Zhenhua and Xu, Jun},
  booktitle={Proceedings of the 30th ACM SIGKDD Conference on Knowledge Discovery and Data Mining},
  pages={6437--6447},
  year={2024}
}

@inproceedings{annamoradnejad2019comprehensive,
  title={A comprehensive analysis of twitter trending topics},
  author={Annamoradnejad, Issa and Habibi, Jafar},
  booktitle={2019 5th international conference on web research (ICWR)},
  pages={22--27},
  year={2019},
  organization={IEEE}
}

@article{jun2018ten,
  title={Ten years of research change using Google Trends: From the perspective of big data utilizations and applications},
  author={Jun, Seung-Pyo and Yoo, Hyoung Sun and Choi, San},
  journal={Technological forecasting and social change},
  volume={130},
  pages={69--87},
  year={2018},
  publisher={Elsevier}
}

@inproceedings{tan2025htmlrag,
  title={Htmlrag: Html is better than plain text for modeling retrieved knowledge in rag systems},
  author={Tan, Jiejun and Dou, Zhicheng and Wang, Wen and Wang, Mang and Chen, Weipeng and Wen, Ji-Rong},
  booktitle={Proceedings of the ACM on Web Conference 2025},
  pages={1733--1746},
  year={2025}
}

@techreport{kincaid1975derivation,
  title={Derivation of new readability formulas (automated readability index, fog count and flesch reading ease formula) for navy enlisted personnel},
  author={Kincaid, J Peter and Fishburne Jr, Robert P and Rogers, Richard L and Chissom, Brad S},
  year={1975}
}

@article{flesch2007flesch,
  title={Flesch-Kincaid readability test},
  author={Flesch, Rudolf},
  journal={Retrieved October},
  volume={26},
  number={3},
  pages={2007},
  year={2007}
}

@inproceedings{shap,
  title={A Unified Approach to Interpreting Model Predictions},
  author={Lundberg, Scott M. and Lee, Su-In},
  booktitle={Advances in Neural Information Processing Systems (NeurIPS)},
  pages={4765--4774},
  year={2017}
}

@inproceedings{ding2025citations,
  title={Citations and trust in llm generated responses},
  author={Ding, Yifan and Facciani, Matthew and Joyce, Ellen and Poudel, Amrit and Bhattacharya, Sanmitra and Veeramani, Balaji and Aguinaga, Sal and Weninger, Tim},
  booktitle={Proceedings of the AAAI Conference on Artificial Intelligence},
  volume={39},
  number={22},
  pages={23787--23795},
  year={2025}
}

@article{chen2024humans,
  title={Humans or llms as the judge? a study on judgement biases},
  author={Chen, Guiming Hardy and Chen, Shunian and Liu, Ziche and Jiang, Feng and Wang, Benyou},
  journal={arXiv preprint arXiv:2402.10669},
  year={2024}
}

@article{diaz2022echo,
  title={Echo chambers and information transmission biases in homophilic and heterophilic networks},
  author={Diaz-Diaz, Fernando and San Miguel, Maxi and Meloni, Sandro},
  journal={Scientific Reports},
  volume={12},
  number={1},
  pages={9350},
  year={2022},
  publisher={Nature Publishing Group UK London}
}

@misc{llama4scout,
  author = {{Meta AI}},
  title = {Everything we announced at our first-ever LlamaCon},
  year = {2025},
  month = {April},
  howpublished = {\url{https://ai.meta.com/blog/llamacon-llama-news/}},
  note = {Accessed: 2025-11-27. Announces Llama 4 Scout (109B MoE, 17B active).}
}

@inproceedings{yang-etal-2018-hotpotqa,
    title = "{H}otpot{QA}: A Dataset for Diverse, Explainable Multi-hop Question Answering",
    author = "Yang, Zhilin  and
      Qi, Peng  and
      Zhang, Saizheng  and
      Bengio, Yoshua  and
      Cohen, William  and
      Salakhutdinov, Ruslan  and
      Manning, Christopher D.",
    editor = "Riloff, Ellen  and
      Chiang, David  and
      Hockenmaier, Julia  and
      Tsujii, Jun{'}ichi",
    booktitle = "Proceedings of the 2018 Conference on Empirical Methods in Natural Language Processing",
    month = oct # "-" # nov,
    year = "2018",
    address = "Brussels, Belgium",
    publisher = "Association for Computational Linguistics",
    url = "https://aclanthology.org/D18-1259/",
    doi = "10.18653/v1/D18-1259",
    pages = "2369--2380"
}

@misc{llama4maverick,
  author = {{Meta AI}},
  title = {Everything we announced at our first-ever LlamaCon},
  year = {2025},
  month = {April},
  howpublished = {\url{https://ai.meta.com/blog/llamacon-llama-news/}},
  note = {Accessed: 2025-11-27. Announces Llama 4 Maverick (400B MoE, 17B active).}
}

@misc{qwen3dense,
  author = {{Alibaba Cloud Qwen Team}},
  title = {Qwen3: Think Deeper, Act Faster},
  year = {2025},
  month = {April},
  howpublished = {\url{https://qwenlm.github.io/blog/qwen3/}},
  note = {Accessed: 2025-11-27. Release of Qwen 3 dense models, including 32B variant.}
}

@misc{deepseekv3,
  author = {{DeepSeek AI}},
  title = {DeepSeek-V3 Technical Report},
  year = {2024},
  month = {December},
  howpublished = {\url{https://github.com/deepseek-ai/DeepSeek-V3/blob/main/DeepSeek_V3.pdf}},
  note = {Accessed: 2025-11-27. Details 671B MoE model with 37B active parameters.}
}

@article{landis1977measurement,
  title={The measurement of observer agreement for categorical data},
  author={Landis, J Richard and Koch, Gary G},
  journal={biometrics},
  pages={159--174},
  year={1977},
  publisher={JSTOR}
}

@article{poudel2025social,
  title={Social and political framing in search engine results},
  author={Poudel, Amrit and Weninger, Tim},
  journal={arXiv preprint arXiv:2507.13325},
  year={2025}
}

@inproceedings{wang2016learning,
  title={Learning to rank with selection bias in personal search},
  author={Wang, Xuanhui and Bendersky, Michael and Metzler, Donald and Najork, Marc},
  booktitle={Proceedings of the 39th International ACM SIGIR conference on Research and Development in Information Retrieval},
  pages={115--124},
  year={2016}
}

@inproceedings{deng2025cram,
  title={Cram: Credibility-aware attention modification in llms for combating misinformation in rag},
  author={Deng, Boyi and Wang, Wenjie and Zhu, Fengbin and Wang, Qifan and Feng, Fuli},
  booktitle={Proceedings of the AAAI Conference on Artificial Intelligence},
  volume={39},
  number={22},
  pages={23760--23768},
  year={2025}
}

@inproceedings{yang2025accuracy,
  title={Accuracy and political bias of news source credibility ratings by large language models},
  author={Yang, Kai-Cheng and Menczer, Filippo},
  booktitle={Proceedings of the 17th ACM Web Science Conference 2025},
  pages={127--137},
  year={2025}
}

@misc{dean2025ctr,
  author       = {Dean, Brian},
  title        = {We Analyzed 4 Million Google Search Results. Here's What We Learned About Organic Click Through Rate},
  howpublished = {\url{https://backlinko.com/google-ctr-stats}},
  note         = {Backlinko},
  year         = {2025},
  month        = {April 16}
}

@article{xu2000inference,
  title={Inference for generalized Gini indices using the iterated-bootstrap method},
  author={Xu, Kuan},
  journal={Journal of Business \& Economic Statistics},
  volume={18},
  number={2},
  pages={223--227},
  year={2000},
  publisher={Taylor \& Francis}
}

@article{mckight2010kruskal,
  title={Kruskal-wallis test},
  author={McKight, Patrick E and Najab, Julius},
  journal={The corsini encyclopedia of psychology},
  pages={1--1},
  year={2010},
  publisher={Wiley Online Library}
}
\clearpage

\begin{appendices}
\crefalias{section}{appendix}
\section{Ethics}
\label{app:ethics}
To the best of our knowledge, our dataset contains no personal identifiers or information.
To mitigate risks of data misuse, we collect only publicly accessible information and follow ethical guidelines for social data research~\cite{2012-dittrich-mraf}.  
Our data collection pipeline incorporates rate limiting, exponential backoff, and concurrency controls to avoid impacting the performance of search engines.  
Furthermore, a waiver was obtained from the institutional ethics committee.



\section{Bootstrapped Evaluation of the Gini Index}
\label{app:gini}

To assess differences in inequality across \acp{aise} and \acp{tse}, we employed a bootstrap procedure to evaluate Gini indices. For each iteration, we resampled queries with replacement from the observed set of search words for each system, 
constructed domain count distributions, and computed the corresponding Gini index. 
The difference between the 
two indices was recorded for that replicate. Repeating this process 1,000 times yielded an empirical distribution 
of Gini differences. From this distribution, we derived a 95\% confidence interval by taking the 2.5th and 97.5th 
percentiles. We also computed the observed difference using the full domain counts, which served as a baseline 
for comparison. This non‑parametric approach allows us to quantify the robustness of observed differences in 
domain inequality, mitigating concerns that results may be driven by sampling variability rather than systematic 
differences between systems. Table~\ref{tab:gini_results} reports the statistics results.

\begin{table}[ht]
\centering
\caption{Two-sided bootstrap $p$-values for Gini index differences between \acp{aise} and \acp{tse}. 
Note: * $p<0.05$, ** $p<0.01$, *** $p<0.001$.}
\begin{tabular}{l l}
\hline
System Pair & $p_{\text{two-sided}}$ \\
\hline
AI Mode vs Bing T.     & 0.59 \\
AI Mode vs Google T.   & 0.58 \\
ChatGPT vs Bing T.     & 0.103 \\
ChatGPT vs Google T.   & 0.026 * \\
Copilot vs Bing T.     & 0.999 \\
Copilot vs Google T.   & 0.999 \\
Gemini vs Bing T.      & 0.000 *** \\
Gemini vs Google T.    & 0.000 *** \\
Grok vs Bing T.        & 0.000 *** \\
Grok vs Google T.      & 0.000 *** \\
Perplexity vs Bing T.  & 0.759 \\
Perplexity vs Google T.& 0.672 \\
\hline
\end{tabular}
\label{tab:gini_results}
\end{table}

\section{Data Annotation} 
\label{app:annotation_results}

\paragraph{Coverage Rate}
\Cref{tab:annotation_result} illustrates the coverage rate of each annotation method shown in the main body of our experiments. 
We evaluate the extent to which each method successfully retrieves categorization data from four distinct external knowledge bases: Forcepoint, Tranco Ranking, \ac{mbfc} and VirusTotal.

\begin{table}[b]
\caption{Percentages of matched domains across annotations.}
\resizebox{\columnwidth}{!}{
\begin{tabular}{@{}lllll@{}}
\toprule
\textbf{} & \textbf{Forcepoint} & \textbf{Tranco Ranking} & \textbf{MBFC} & \textbf{VirusTotal} \\ \midrule
\textbf{AI Mode}     & 90\% & 50\% &17\% &100\% \\
\textbf{ChatGPT}     & 88\% & 57\% &29\% &100\% \\
\textbf{Copilot}     & 78\% & 47\% &16\% &100\% \\
\textbf{Gemini}      & 90\% & 54\% &20\% &100\% \\
\textbf{Grok}        & 85\% & 45\% &17\% &100\% \\
\textbf{Perplexity}  & 89\% & 50\% &17\% &100\% \\
\textbf{Google}      & 89\% & 51\% &18\% &100\% \\
\textbf{Bing}        & 84\% & 54\% & 9\% &100\% \\ \bottomrule
\end{tabular}}
\label{tab:annotation_result}
\end{table}

\paragraph{Validations}
To validate the correctness of existing annotations, we apply a multi-stage human verification to the Forcepoint and VirusTotal labels.  
Note, we do not perform validation for Tranco or \ac{mbfc} because neither dataset provides semantic or credibility labels suitable for our verification protocol. 
Tranco is a domain-ranking list aggregated from multiple traffic–estimation sources; it contains no human- or model-interpretable annotations, only popularity scores. 
Since our validation procedure is designed to assess label correctness, Tranco falls outside the scope of label-level verification. 
Similarly, we omit validation for \ac{mbfc}, as its assessments are produced directly by trained human reviewers who evaluate outlets based on fact-checking history, content analysis, and editorial practices. 

Three co-authors independently annotate a randomly selected subset of 100 samples from each (\ie Forcepoint and VirusTotal) dataset, while remaining blinded to the original labels. 
We adopt the same categorization taxonomy as the original third-party annotations to ensure direct comparability.
Ground-truth labels are then established using a majority-voting scheme: a label is adopted only if it receives agreement from at least two annotators. 
Instances that fail to meet this threshold are treated as ambiguous and are subsequently resolved through deliberative group discussion until a final classification is reached.
We then calculated Fleiss' $\kappa$ to measure the agreement between the human-verified labels and the original third-party annotations. 
The resulting scores of Forcepoint and Virustotal reach 92\% and 86\%, respectively.
These scores ($> 80\%$) indicate a high level of consistency between our manual audit and the external datasets~\cite{landis1977measurement}. 
This statistical evidence confirms that the third-party annotations align with human judgment, thereby validating their accuracy for use in our measurements.


\section{Overlapped Domains with \acp{tse}}
\label{app:coverage_ratio}

\begin{table}[t]
\centering
\caption{Mean domain overlap between \acp{aise} and traditional search engines (TSEs).}
\begin{tabular}{lcc}
\toprule
\textbf{Search Engine} & \textbf{Bing (Mean)} & \textbf{Google (Mean)} \\
\midrule
ChatGPT     & 0.041 & 0.118 \\
Copilot     & 0.078 & 0.178 \\
Gemini      & 0.022 & 0.120 \\
Grok        & 0.014 & 0.055 \\
Perplexity  & 0.081 & 0.408 \\
\bottomrule
\end{tabular}
\label{tab:domain_overlap}
\end{table}

\Cref{tab:domain_overlap} presents the mean domain overlap between \acp{aise} and \acp{tse}.
Values indicate the per response ratio of overlapping domains divided by the number of domains retrieved by Google or Bing.


\section{Rank Turbulence Divergence (RTD) }
\label{app:rtd}
We compute an RTD score for each domain, with values bounded between $0$ and $1$.  
In all cases, $r_{\tau,1}$ denotes the rank derived from \acp{tse}, while $r_{\tau,2}$ represents the rank obtained from \acp{aise}.  
To capture directional bias, we adopt a signed formulation: RTD values approaching $+1$ indicate that the domain is more prominently ranked in \acp{tse} results, whereas values approaching $-1$ signify stronger representation within \acp{aise} outputs.  
We count the RTD score pair-by-pair between \acp{tse} and \acp{aise}.
The RTD calculates the element-wise divergence as follows:
\begin{equation}
\left| \frac{1}{r_{\tau,1}} - \frac{1}{r_{\tau,2}} \right|
\end{equation}
Here, $\tau$ denotes a domain, and $r_{\tau,1}$ and $r_{\tau,2}$ represent its ranks in lists $R_1$ and $R_2$, respectively. 
If a domain appears in only one of the lists, we assign it a rank of $\max(R) + 1$ in the other list to ensure consistent comparison across both corpora.
This formulation introduces a bias toward higher-ranked tokens. 
To control this bias, we introduce a parameter $\alpha$:
\begin{equation}
\left| \frac{1}{r_{\tau,1}^{\alpha}} - \frac{1}{r_{\tau,2}^{\alpha}} \right|^{\frac{1}{\alpha + 1}}
\end{equation}
For each domain in the union of $R_1$ and $R_2$, divergence is computed using Equation (2). 
We set $\alpha = \frac{1}{3}$, which has been shown to yield a balanced representation~\cite{dodds2023allotaxonometry,poudel2024navigating}.



\section{MBFC Factual Score Transformation}
\label{app:mbfc_transform}
The original factual scores from \acp{mbfc} range from $-1$ (highly factual) to $10$ (lowly factual), which is a nonstandard orientation. 
To align with conventional political score normalization (ranging from $-10$ to $10$), we apply a linear transformation as follows:
\begin{equation}
    \text{Factual Score}_{\mathrm{new}} = 10 - \frac{20(\text{Factual Score}_{\mathrm{MBFC}} + 1)}{11}
\end{equation}
where $\text{Factual Score}_{\mathrm{MBFC}}$ is the original MBFC factual score. After transformation, the scale is standardized such that $-10$ denotes the lowest factuality and $10$ denotes the highest factuality.

\section{Testing the Independence of the MBFC Factual Score} \label{app:test_cp}

\begin{figure}[t]
    \vspace{-2ex} 
    \centering
    \begin{subfigure}[b]{0.48\columnwidth}
        \centering        \includegraphics[width=\textwidth]{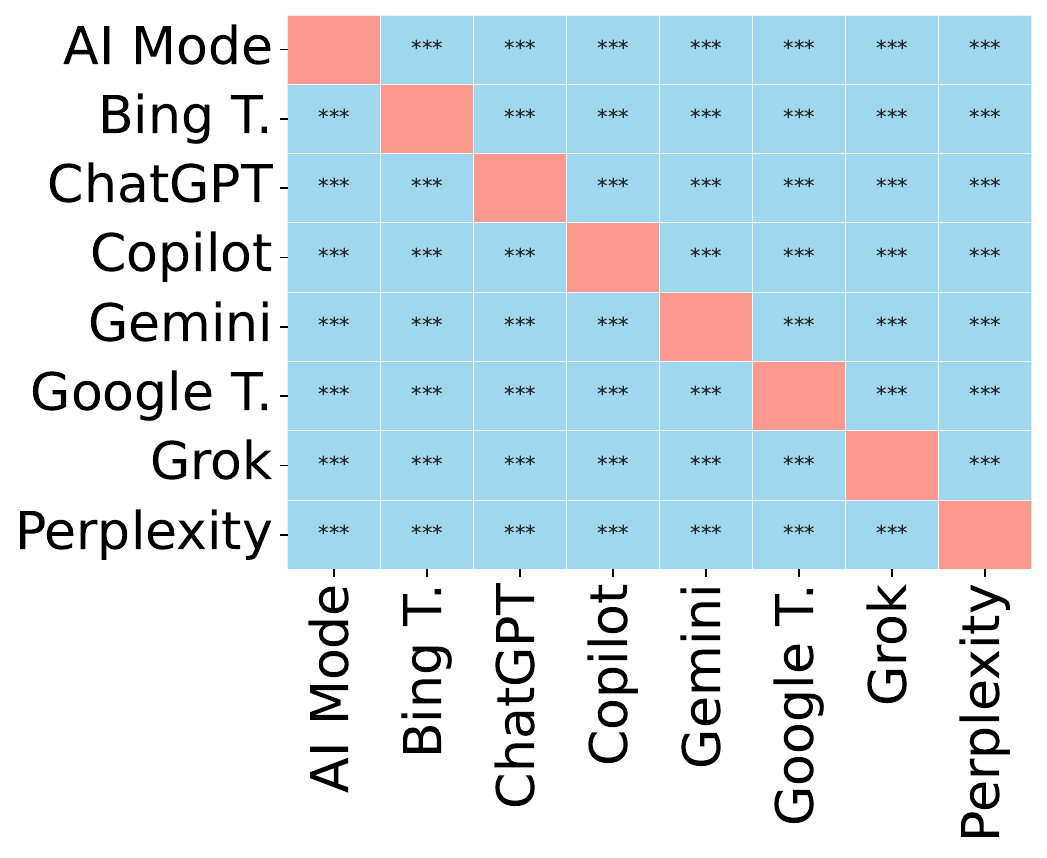}
        \caption{}
        \label{fig:post_hoc_factual}
    \end{subfigure}
    \hfill
    \begin{subfigure}[b]{0.48\columnwidth}
        \centering
    \includegraphics[width=\textwidth]{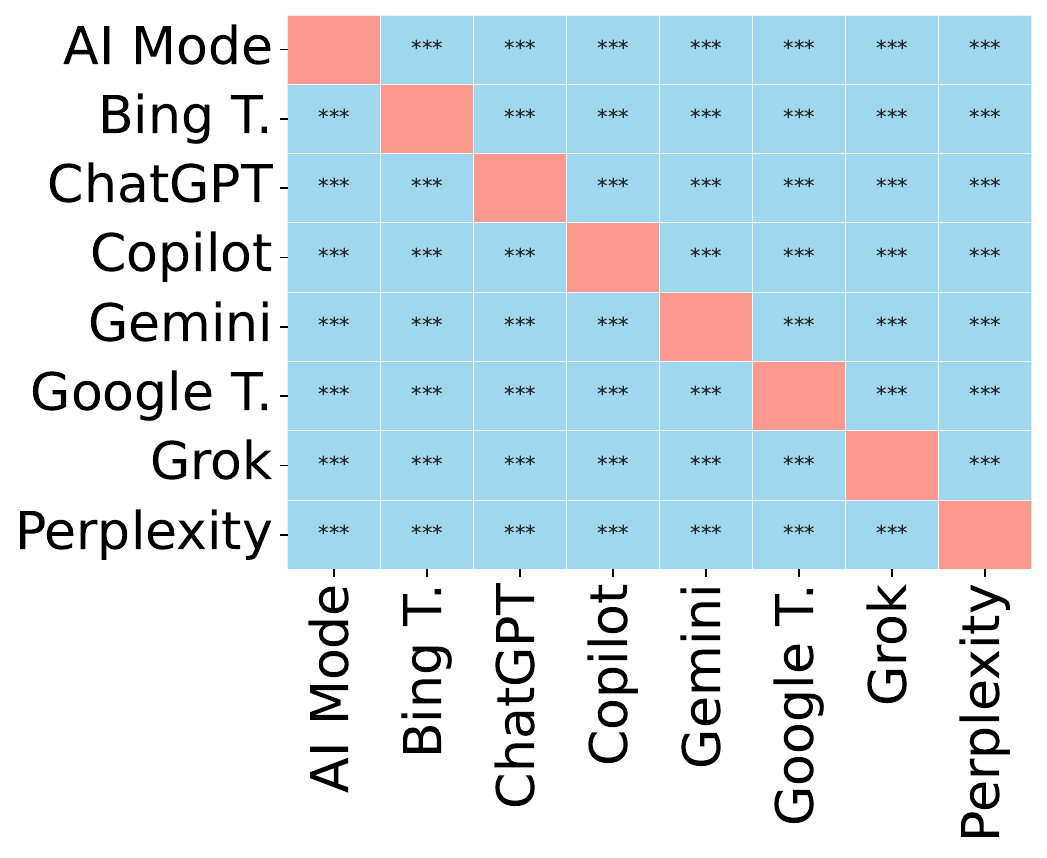}
        \caption{}
        \label{fig:post_hoc_bias}
    \end{subfigure}
    \caption{Heatmap for Dunn's Post-hoc Test (Bonferroni-adjusted p-values) (a) for credibility and (b) for political. All pairwise comparisons are statistically significant. }
    \label{fig:post_hoc}
\end{figure}

We conducted non-parametric independence tests to examine whether MBFC factual reporting scores, credibility, 
and political leaning are statistically associated with different search engines. Specifically, we applied the 
Kruskal–Wallis (KW) test to compare distributions across groups.

For each variable (credibility $H=1079054$,$p<0.001$ and political $H=1943828$, $p<0.001$), the KW test indicated significant differences 
among search engines. To identify which pairs of search engines differed, we performed 
post-hoc pairwise comparisons using Dunn’s test with Bonferroni correction. Figure~\ref{fig:post_hoc} reports the Dunn’s test results on (a)credibility and (b)political. The KW and post-hoc results suggest that search engines differ systematically in the MBFC 
profiles of the domains they surface, with AISEs more often associated with higher factual reporting scores 
and credibility, and some exhibiting distinct political lean distributions.

\section{Threats Threshold}
\label{app:virus_threshold}



\begin{figure}
    \centering
    \includegraphics[width=\linewidth]{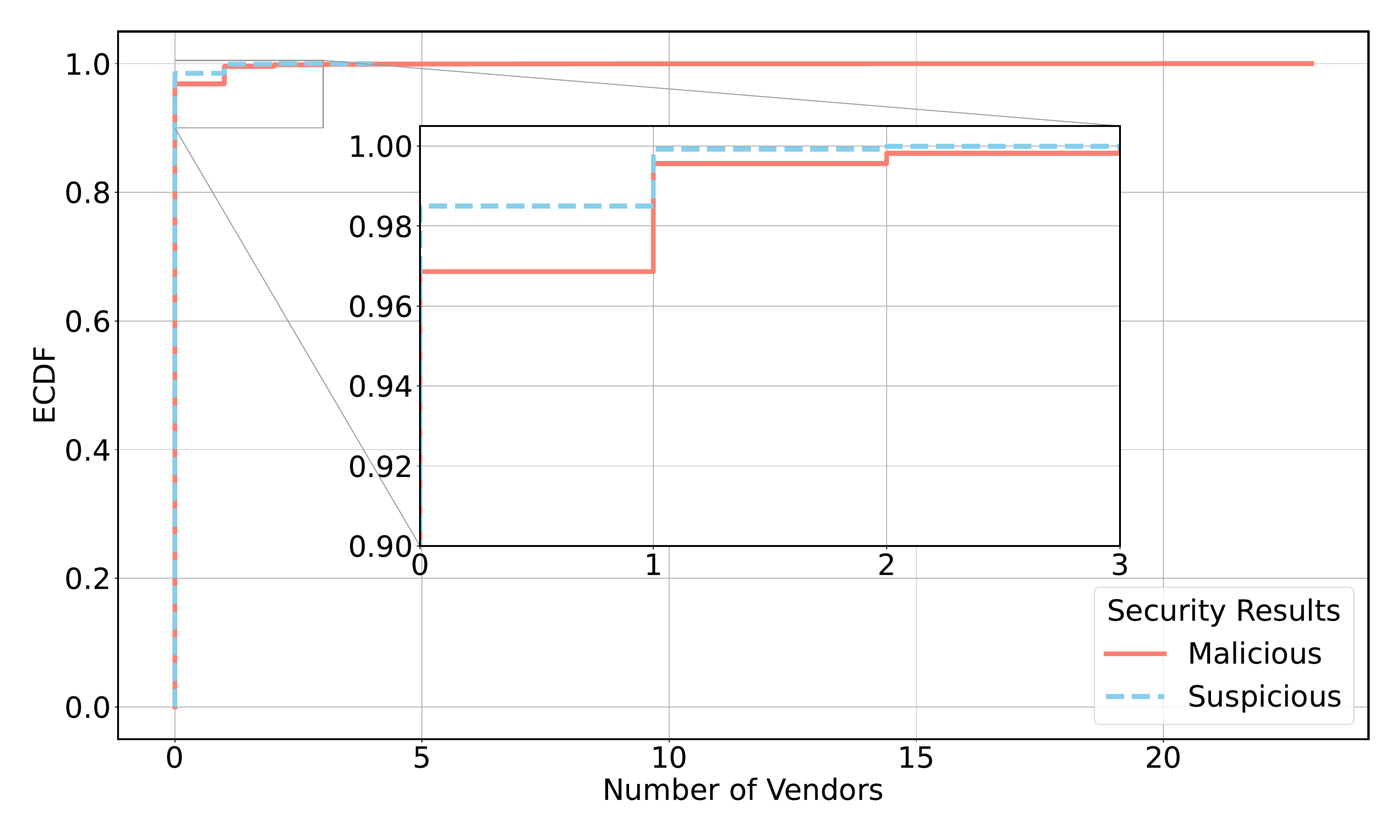}
    \caption{ECDF of domains by number of security vendors reporting}
    \label{fig:virus_number}
\end{figure}

To determine the detection threshold in \S\ref{rq2_cyber_threats}, we analyze how many security vendors label each domain in our dataset as malicious or suspicious.  
\Cref{fig:virus_number} shows the empirical cumulative distribution function (ECDF) of vendor tags across all domains.  
Overall, 3.1\% (4,414) of domains are flagged as malicious and 1.2\% (2,075) as suspicious by more than one vendor.  
However, the number of detected malicious domains decreases sharply as the threshold increases (Threshold 2: 0.4\% or 646 domains, Threshold 3: 0.1\% or 273 domains).  
Though a threshold of one positive flag maximizes sensitivity for emerging threats, it risks high false positives due to vendor inconsistencies~\cite{imc_virus_total_threshold}.  
To balance false positives and false negatives, we adopt a stricter threshold of at least two vendor detections, classifying 0.4\% (646) of domains as malicious, in line with established practices~\cite{10.1145/3673660.3655042,imc_virus_total_threshold}.  

\section{Query Categorization}
\label{app:query_categorization}
\paragraph{Methodology}
We utilize five open-source \acp{llm} to categorize the queries. 
\Cref{tab:model_selection} summarizes the model configurations. 
Each model independently annotates all queries. 
To ensure direct comparability, we adopt Google's official category taxonomy~\cite{google_trend}. 
Ground-truth labels are determined via majority voting: a label is accepted only if at least three models agree. 
Queries that do not meet this threshold are marked as ambiguous and subsequently resolved through human annotation from one of the authors. 
For each labeling task, we retrieve the complete set of categories and subcategories from Google's taxonomy JSON,\footnote{\url{https://trends.google.com/trends/api/explore/pickers/category?hl=en-US}} 
parse all 25 top-level categories and their subcategories, and construct the full category list used in our annotation pipeline.

\begin{table}[t]
\centering
\caption{LLMs used for categorizing the queries}
\begin{tabular}{lccc}
\toprule
\textbf{Model Family} & \textbf{Release} & \textbf{Parameters} & \textbf{Arch.} \\
\midrule
Llama~4 Scout~\cite{llama4scout}       & 2025.04  & 109B total/17B active  & MoE  \\
Llama~4 Maverick~\cite{llama4maverick} & 2025.04  & 400B total/17B active  & MoE  \\
Qwen~3~\cite{qwen3dense}               & 2025.04  & 4B                     & Dense \\
Qwen~3~\cite{qwen3dense}               & 2025.04  & 14B                    & Dense \\
DeepSeek-V3~\cite{deepseekv3}          & 2024.12  & 671B total/37B active  & MoE  \\
\bottomrule
\end{tabular}
\label{tab:model_selection}
\end{table}

\begin{tcolorbox}[
    colback=pink!30,        
    colframe=pink!70!black, 
    boxrule=1pt,
    arc=3pt,
    boxsep=3pt,
    left=3pt, right=3pt, top=3pt, bottom=3pt
]
1. Arts \& Entertainment (Celebrities \& Entertainment News, Comics \& Animation, ...)

2. Autos \& Vehicles (...)

...
\end{tcolorbox}

We then replace ``\{categories\_text\}'' with it in the following prompt, to label the query with 10 tries:

\begin{tcolorbox}[
    colback=pink!30,        
    colframe=pink!70!black, 
    boxrule=1pt,
    arc=3pt,
    boxsep=3pt,
    breakable,
    left=3pt, right=3pt, top=3pt, bottom=3pt
]
\textbf{Prompt}: 
You are a professional theme classification assistant. Your task is to analyze the given text and classify it into the most relevant categories from the provided list.
\\

AVAILABLE CATEGORIES:
\{categories\_text\}
\\

CLASSIFICATION RULES:

1. Carefully analyze the text's main themes, content, and keywords

2. Identify ALL relevant main categories (1-3 categories maximum)

3. For each main category, select the MOST RELEVANT subcategory

4. Rank categories by relevance (most relevant first)

5. If text covers multiple topics, include all relevant categories

6. If no exact subcategory match exists, choose the CLOSEST one

7. Consider the text's context, intent, and all purposes

8. If the text doesn't clearly fit any of the provided categories, output Unknown
\\

IMPORTANT OUTPUT FORMAT:
\\

You MUST respond with EXACTLY this format (no additional text, no explanations):
\\

For 1 category: \{MAIN\_CATEGORY\}\{SUBCATEGORY\}
\\

For 2 categories: \{MAIN\_CATEGORY1\}\{SUBCATEGORY1\}| \\ \{MAIN\_CATEGORY2\}\{SUBCATEGORY2\}
\\

For 3 categories: \{MAIN\_CATEGORY1\}\{SUBCATEGORY1\}| \\ \{MAIN\_CATEGORY2\}\{SUBCATEGORY2\}| \\ \{MAIN\_CATEGORY3\}\{SUBCATEGORY3\}
\\

For unknown content: \{Unknown\}\{Unknown\}
\\

Examples of correct output:
\\

\{Arts \& Entertainment\}\{Animated Films\}

\{Health\}\{Alzheimer's Disease\}|\{People \& Society\}\{Family \& Relationships\}

\{Sports\}\{Volleyball\}|\{Health\}\{Fitness\}|\{Arts \& Entertainment\}\{Movies\}

\{Unknown\}\{Unknown\}
\\

INPUT TEXT TO CLASSIFY:
\{text\}
\\

CLASSIFICATION RESULT:
\end{tcolorbox}

\paragraph{Validations}
We follow the validation methodology described in \Cref{app:annotation_results}, excluding categories where \acp{llm} fail to reach consensus (\ie in such cases, the query categories are annotated by one of the authors). 
From the categorization dataset, we randomly sample 100 instances and compute Fleiss'~$\kappa$ to quantify agreement between the human-verified labels and the original third-party annotations.
The resulting $\kappa = 0.88$ reflects a high level of agreement ($> 0.80$)~\cite{landis1977measurement}, indicating that the \ac{llm}-generated labels are strongly aligned with human judgment and suitable for use in our measurements.

\section{Hyperparameter}
\label{app:hyper}
\Cref{tab:hyper_param} illustrates the parameters for our experiments in \S\ref{sec:rq3}.

\begin{table*}[t]
\centering
\caption{Overview of machine learning models, hyperparameter search space, and best-performing configurations.}
\resizebox{\textwidth}{!}{
\begin{tabular}{lll}
\toprule
\textbf{Model} & \textbf{Hyperparameter Grid} & \textbf{Best Parameters} \\
\midrule
Logistic Regression & 
C: \{0.1, 1, 10\}; Penalty: \{l2\} & 
C=10; Penalty=l2; Solver=lbfgs \\

Random Forest & 
n\_estimators: \{100, 200\}; max\_depth: \{None, 10, 20\} & 
n\_estimators=200; max\_depth=None \\

XGBoost & 
n\_estimators: \{100, 200\}; max\_depth: \{3, 6\}; learning\_rate: \{0.05, 0.1\} & 
n\_estimators=200; max\_depth=6; learning\_rate=0.1 \\

KNN & 
n\_neighbors: \{3, 5, 7\} & 
n\_neighbors=3 \\

MLP Neural Net & 
hidden\_layer\_sizes: \{(50,), (100,)\}; alpha: \{0.0001, 0.001\} & 
hidden\_layer\_sizes=(100,); alpha=0.001 \\
\bottomrule
\end{tabular}}
\label{tab:hyper_param}
\end{table*}

\section{Susceptible Query Category}
\label{app:sus_query_cate}
\begin{figure}
    \centering
    \includegraphics[width=\linewidth]{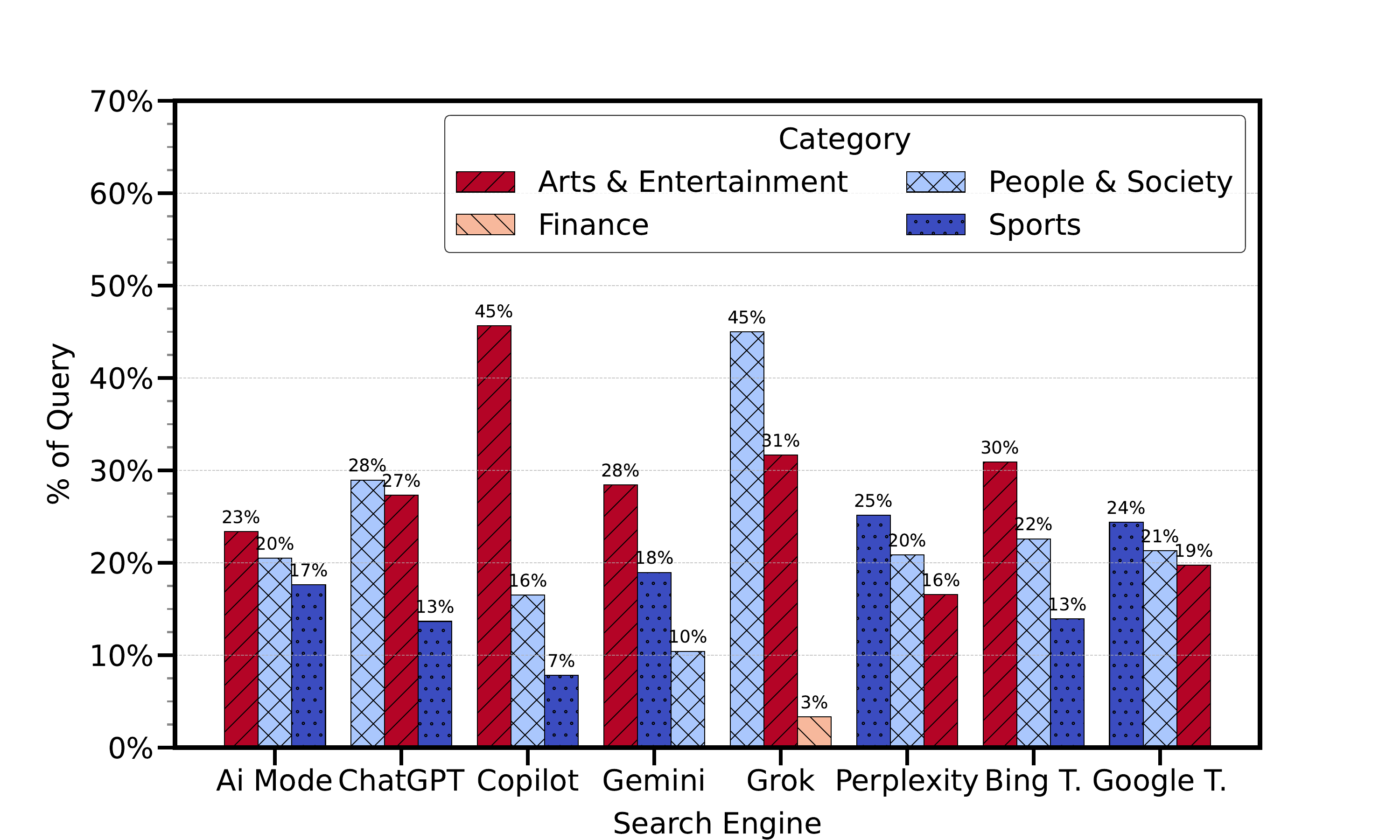}
    \caption{Distribution of Susceptible Query Categories by Frequency}
    \label{fig:top3_malicious_categories_by_engine}
\end{figure}

We adopt the category taxonomy from Google Trends to classify each query (see \Cref{app:query_categorization}).  
\Cref{fig:top3_malicious_categories_by_engine} presents the three most frequent Google Trend categories associated with queries that trigger malicious domains, ranked by their occurrence within each search engine.  

\section{HTML Structures}
\label{app:html_structure}
\begin{itemize}
    \item \textbf{Semantic Tags:} This feature measures the total number of semantic HTML elements, including \texttt{<header>}, \texttt{<nav>}, \texttt{<main>}, \texttt{<article>}, \texttt{<section>}, and \texttt{<footer>}. A higher count reflects adherence to modern web development standards that emphasize meaningful structure, improved document hierarchy, and enhanced accessibility.

    \item \textbf{Nesting Depth:} The nesting depth quantifies the maximum hierarchical level of embedded HTML elements within the Document Object Model (DOM). It captures the structural complexity and organizational depth of a webpage. Greater depth often indicates more elaborate design and layout hierarchy, whereas shallower structures may reflect simpler content organization.

    \item \textbf{Accessibility Features:} This feature represents the number of accessibility-related attributes implemented in the HTML, including \texttt{alt}, \texttt{role}, \texttt{aria-label}, and \texttt{aria-hidden}. These attributes are essential indicators of a page’s accessibility compliance and its accommodation of assistive technologies.

    \item \textbf{Markup Errors:} The number of deprecated or discouraged HTML tags, such as \texttt{<font>} and \texttt{<center>}, is counted as an indicator of poor or outdated coding practices. A higher frequency of such tags implies lower technical quality and weaker adherence to contemporary web standards.
\end{itemize}

\end{appendices}
\end{document}